\documentclass[lettersize,journal]{IEEEtran}
\usepackage{amsmath,amsfonts}
\usepackage{algorithmic}
\usepackage{algorithm}
\usepackage{array}
\usepackage[caption=false,font=normalsize,labelfont=sf,textfont=sf]{subfig}
\usepackage{textcomp}
\usepackage{stfloats}
\usepackage{url}
\usepackage{verbatim}
\usepackage{graphicx}
\usepackage{cite}
\usepackage{multirow}
\usepackage[table]{xcolor} 
\usepackage{tabularx}      
\usepackage{array}         
\usepackage{booktabs}      
\usepackage{pifont}        
\usepackage{threeparttable} 
\usepackage{rotating}      
\usepackage[hidelinks,bookmarks=false]{hyperref}
\newcommand{\best}[1]{\textbf{#1}}
\newcommand{\sbest}[1]{\underline{#1}}

\newcommand{\atenotes}{%
  \begin{tablenotes}[flushleft]\footnotesize
    \item 1) AL=ALIKED, Ra=RaCo, SP=SuperPoint, XF=XFeat. 
    \item 2) LK=optical-flow tracking, LG=LightGlue matching.
    \item 3) \best{Bold}/\sbest{underline} represents best/second per column.
    \item 4) Odometry and loop-closed trajectory are ranked separately.
    \item 5) \ding{55} Tracking failure. 
    \item 6) $^{\dagger}$ Mean over a partial subset. 
    \item 7) $^{*}$ Loop-closed trajectory is worse than odometry.
  \end{tablenotes}}
\newcommand{\atenotesref}{%
  \begin{tablenotes}[flushleft]\footnotesize
    \item Cell format (odometry\,/\,loop-closed), ranking (\best{best}/\sbest{second}), abbreviations, and symbols (\ding{55},\,$^{\dagger}$,\,$^{*}$) as defined in Table~\ref{tab:ate_summary}.
  \end{tablenotes}}

\begin{document}
\bstctlcite{IEEEexample:BSTcontrol}

\title{DL-VINS-Factory: A Modular Framework for Learned Visual Front-Ends in Visual-Inertial SLAM}


\author{Shoon Kit Lim$^{1,2}$,
        Melissa Jia Ying Chong$^{2}$,
        and~Ting Yang Ling$^{3}$%
\thanks{$^{1}$Corresponding Author
(e-mail: shoonkitlim@gmail.com).}
\thanks{$^{2}$S.~K.~Lim and M.~J.~Y.~Chong are with the
University of Southampton Malaysia, Iskandar Puteri, Johor, Malaysia
(e-mail: \{skl1g14, m.j.y.chong\}@soton.ac.uk).}%
\thanks{$^{3}$T.~Y.~Ling is with the University of Southampton,
Southampton SO17 1BJ, U.K.
(e-mail: ivan.ling@soton.ac.uk).}
}

\markboth{Journal of \LaTeX }%
{DL-VINS-Factory: A Comprehensive Framework for Learned Visual Front-Ends in Visual-Inertial SLAM}

\maketitle

\begin{abstract}
Deep-learning features excel in visual matching, yet their practical value in tightly coupled visual-inertial SLAM (VI-SLAM) remains insufficiently characterized. We present DL-VINS-Factory, a unified framework that integrates learned feature extractors (ALIKED, RaCo, SuperPoint, XFeat) with either Lucas--Kanade (LK) optical-flow tracking or LightGlue (LG) descriptor matching. All front-ends share a sliding-window Ceres back-end, with optional AnyLoc DINOv2-VLAD loop closure, and 4-DoF pose-graph optimization. We benchmark the system across the four datasets covering indoor, unstructured outdoor, aggressive-motion, and visually degraded conditions. Results show that learned front-ends are viable for real-time embedded VI-SLAM, but are not universally superior to classical tracking. Relative to the corresponding GFTT+LK baseline, ALIKED+LG reduces EuRoC ATE by $5\%$ in monocular odometry and by $7\%$ in stereo with loop-closure. On NTU-VIRAL, where aggressive aerial motion increases inter-frame viewpoint change, ALIKED+LG stereo reduces loop-closed ATE by $12\%$. In Botanic Garden dataset, optical-flow tracking remains preferable, but learned keypoints still improve over the baseline GFTT, in which SuperPoint+LK reduces grayscale camera ATE by $29\%$, while RaCo+LK reduces RGB camera ATE by $38\%$. On SubT-MRS, learned front-ends display varying degree of improvement based on individual cases. With TensorRT acceleration on a Jetson AGX Orin, all valid configurations run in real time between $29$--$47$ FPS in monocular mode and $18$--$33$ FPS in stereo mode for the EuRoC and NTU-VIRAL datasets. AnyLoc further confirms roughly $2$--$7\times$ more valid loops than BRIEF+DBoW2. The implementation is open-sourced at \url{https://github.com/limshoonkit/DL-VINS-Factory-ROS2/}.
\end{abstract}


\section{Introduction}
\label{sec:intro}

\IEEEPARstart{V}{isual}-Inertial Simultaneous Localization and Mapping (VI-SLAM) is widely used in mobile robotics due to its lightweight, low-cost sensing, but its robustness remains highly dependent on the visual front-end. While active sensors such as LiDAR and radar offer complementary sensing capabilities, camera-based SLAM remains attractive due to its rich visual information and favorable cost, weight, and power characteristics. Classical pipelines relying on handcrafted features such as ORB~\cite{orb} or BRISK~\cite{brisk} are efficient but often degrade under low texture, motion blur, and illumination changes ~\cite{benchmark5}. Recent advances in deep learning have introduced learned feature extractors such as SuperPoint \cite{SuperPoint} and, when paired with learned matchers such as SuperGlue~\cite{superglue}, further improve correspondence quality by jointly reasoning over descriptor similarity and spatial context, enabling increasingly robust visual front-ends.

For long-term operation, SLAM systems mitigate accumulated drift through a pipeline consisting of Visual Place Recognition (VPR), geometric loop verification, and Pose Graph Optimization (PGO)~\cite{benchmark1}. Candidate revisits are first retrieved using global image representations before being validated through local feature matching and geometric consistency checks. The resulting loop constraints are then incorporated into a pose graph to enforce global trajectory consistency.

Despite the rapid progress of learned visual front-ends, their integration into complete VI-SLAM systems remains fragmented. Existing studies often evaluate different learned extractors and matchers under varying back-end optimizers, loop-closure strategies, datasets, and hardware platforms, making it difficult to isolate the true impact of the visual front-end. Furthermore, many loop-closure systems remain tightly coupled to descriptor-specific Bag-of-Words (BoW) vocabularies~\cite{dbow2}, requiring additional engineering whenever the underlying local feature representation is replaced. These limitations hinder both fair benchmarking and the practical adoption of emerging learned features.

To address these challenges, we present \textbf{DL-VINS-Factory}, a modular VI-SLAM framework that decouples learned visual front-ends from the estimation back-end and loop-closure pipeline. This design enables controlled comparison of various feature extractors and matcher combinations within a tightly coupled VI-SLAM architecture while supporting front-end agnostic loop retrieval through AnyLoc \cite{AnyLoc}.

The contributions of this work are summarized as follows:

\begin{itemize}

\item We develop a modular VI-SLAM framework that enables interchangeable learned visual front-ends, including ALIKED, RaCo, SuperPoint, and XFeat, combined with either optical-flow tracking or LightGlue feature matching within a common tightly coupled sliding-window optimization back-end. 

\item We integrate a loop-closure module that performs image-level retrieval using DINOv2 patch embeddings aggregated with a universal Vector of Locally Aggregated Descriptors (VLAD) codebook. This removes the need for per-descriptor BoW vocabulary construction when switching front-ends, while geometric verification reuse the active local feature descriptor through brute-force matching, fundamental-matrix RANSAC, and Perspective-n-Point RANSAC.

\item We conduct a systematic and extensive evaluation across different public datasets covering indoor, outdoor, aggressive-motion, and degraded visual conditions, on both an RTX 3080 Ti workstation and an NVIDIA Jetson AGX Orin. Handcrafted baselines are included as controls under identical experimental conditions to isolate the effect of learned representations.

\end{itemize}

\begin{figure*}[t]
  \centering
  \includegraphics[width=\textwidth]{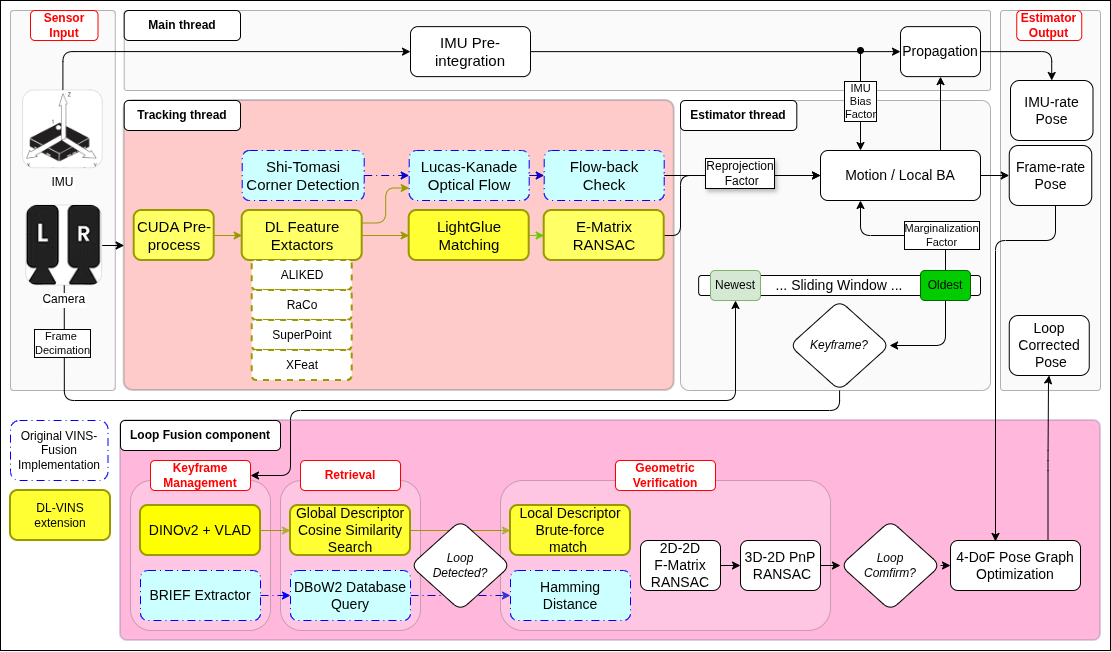}
  \caption{System overview of the proposed DL-VINS-Factory framework. Dashed blue blocks represent the original VINS-Fusion implementation, whereas solid yellow blocks highlight the learned feature front-end extensions and unified loop closure module.}
  \label{fig:system}
\end{figure*}

\section{Related Work}
\label{sec:related}

\begin{table*}[htbp]
\centering
\small 
\renewcommand{\arraystretch}{1.3} 
\caption{Comparison of Traditional and Deep Learning-Based Visual-Inertial SLAM Systems}
\label{tab:slam_comparison}

\begin{tabularx}{\textwidth}{@{} l >{\raggedright\arraybackslash}X >{\raggedright\arraybackslash}X >{\raggedright\arraybackslash}X @{}}
\toprule

\textbf{VI-SLAM System} & \textbf{Frontend (Detection \& Tracking)} & \textbf{Backend} & \textbf{VPR \& Loop Closure} \\

\midrule

\multicolumn{4}{@{}l@{}}{\textbf{Classic}} \\ 
\midrule
ORB-SLAM3 \cite{orbslam3} & ORB features, map projection matching & Co-Visibility Graph
Local BA (Gauss-Newton g2o) & DBoW2 (ORB), PGO, Global BA \\

OKVIS2 \cite{okvis2} & BRISK 2, multi-frame keypoint \& descriptor matching, Fast-SCNN filtering & Bounded factor graph (Ceres) & DBoW2 (BRISK 2), Async full graph opt. \\

VINS-Fusion \cite{vins_fusion} & KLT Tracker & Sliding window Local BA (Ceres) & DBoW2 (BRIEF), 4-DoF PGO \\

OpenVINS \cite{openvins} & KLT Tracker & MSCKF (Sliding window filter) & None built-in, \textit{Extended via VINS-Fusion's loop closure} \\

Kimera \cite{kimera} & KLT Tracker & Fixed-lag smoother (GTSAM / iSAM2) & DBoW2 (ORB), Robust PGO (PCM) \\

\midrule

\multicolumn{4}{@{}l@{}}{\textbf{Deep Learning (DL)}} \\
\midrule
AirSLAM \cite{airslam} & PLNet (unified CNN based on SuperPoint and U-Net backbone), LightGlue keypoint matching, geometric line matching & Local BA (Levenberg-Marquardt g2o) & DBoW2 (PLNet), PGO, Global BA \\

GCN-SLAM \cite{gcnv2} & GCNv2 (ORB style binary descriptors), nearest-neighbor search for matching & \textit{Based on ORB-SLAM2} & DBoW2 (GCNv2), PGO, Global BA \\

DXSLAM \cite{dxslam} & HF-Net (SuperPoint style descriptor), brute force matching & \textit{Based on ORB-SLAM2} & FBoW (SuperPoint), NetVLAD, Global BA \\

HFNet-SLAM \cite{hfnet-slam} & HF-Net (SuperPoint style descriptor), L2-norm descriptors matching & \textit{Based on ORB-SLAM3} & NetVLAD, L2 descriptor similarity score, Global BA \\

SupSLAM \cite{SUPSLAM} & SuperPoint, LK optical flow & \textit{Based on OpenVINS} & \textit{Based on VINS-Fusion} \\

D-VINS \cite{dvins} & SuperPoint, LightGlue matching & \textit{Based on VINS-Fusion} & Faiss (MixVPR global descriptors), SuperPoint-LightGlue geometric verification, 4-DOF PGO \\

SuperVINS \cite{supervins} & SuperPoint, LightGlue matching & \textit{Based on VINS-Fusion} & DBoW3 (SuperPoint), 4-DoF PGO \\

\midrule

\rowcolor{gray!20} 
DL-VINS-Factory (Ours) & Modular extractors (ALIKED, RaCo, SuperPoint, XFeat) with interchangeable LK optical flow or LightGlue matching & \textit{Based on VINS-Fusion} & AnyLoc (DINOv2-VLAD), cosine descriptor similarity, 4-DoF PGO \\

\bottomrule
\end{tabularx}
\end{table*}

\subsection{VI-SLAM with Deep-Learning Feature Extractor}

Widely known VI-SLAM systems such as VINS-Fusion~\cite{vins_fusion}, ORB-SLAM3~\cite{orbslam3}, OKVIS2~\cite{okvis2}, OpenVINS~\cite{openvins}, and Kimera~\cite{kimera} typically couple handcrafted visual front-ends with descriptor-specific back-end or loop-closure vocabularies. Motivated by the limitations of handcrafted features, recent systems have incorporated learned extractors, most notably SuperPoint \cite{SuperPoint}, into established SLAM, as summarized in Table~\ref{tab:slam_comparison}. However, other learned extractors and descriptors remain comparatively under-evaluated. DL-VINS-Factory addresses this gap by providing a common sliding-window optimization back-end and unified evaluation protocol, enabling different front-ends to be compared under identical conditions, which offers more isolated insight on switching front-end.

\subsection{Alternative Deep Learning SLAM Paradigms}

Several deep-learning paradigms address SLAM beyond conventional feature extraction and matching. Detector-free matchers such as LoFTR~\cite{loftr} replace explicit keypoint detection with transformer-based coarse-to-fine correspondence estimation, while dense optical-flow methods such as RAFT~\cite{raft} avoid explicit descriptors altogether. RAFT-VINS~\cite{raft-vins} adapts this idea to VI-SLAM, and DROID-SLAM~\cite{droid_slam} combines flow-based tracking with differentiable bundle adjustment. Sparse recurrent methods such as DPV-SLAM~\cite{dpv-slam} instead track learned patches, improving efficiency over dense deep SLAM while retaining learned iterative optimization. However, these approaches remain challenging for resource-constrained embedded deployment due to transformer, recurrent, or iterative inference, high GPU-memory demand, and limited tight IMU integration~\cite{benchmark5}. Lightweight learned front-ends within conventional VI-SLAM therefore remain a more practical approach for mobile robotics.

\subsection{Visual Place Recognition for Loop Closure}
Conventional VI-SLAM loop closure relies on Bag-of-Words vocabulary~\cite{dbow2} built from local feature descriptors, creating a hard coupling between the visual front-end and the retrieval vocabulary such that changing the front-end requires rebuilding the entire vocabulary. Deep global descriptors decouple retrieval from local feature type. NetVLAD~\cite{netvlad} aggregates features into a holistic representation, while MixVPR~\cite{mixvpr}, EigenPlaces~\cite{eigenplaces}, and CosPlace~\cite{cosplace} exploit multi-layer embeddings and view-robust training paradigms for retrieval under viewpoint and illumination change. Our framework adopts AnyLoc~\cite{AnyLoc}, which aggregates DINOv2~\cite{dinov2} patch embeddings with a single universal VLAD codebook pre-built across diverse environments, achieving descriptor-agnostic retrieval without per-dataset or per-front-end retraining.

\section{System Overview}
\label{sec:system}
Our system is largely built upon VINS-Fusion \cite{vins-mono}\cite{vins_fusion}, and we refer readers to the original literature for foundational details. Our re-implementation adds support for modern compiler threading process, Ceres solver version update, numerical bounds, and updated ROS wrappers, alongside other cosmetic changes. Consequently, the section focuses exclusively on detailing our core differences as depicted in Fig. \ref{fig:system}.

\subsection{Learned Visual Front-end}
\label{sec:learned_front}

\begin{figure*}[t]
  \centering
  \includegraphics[width=\textwidth]{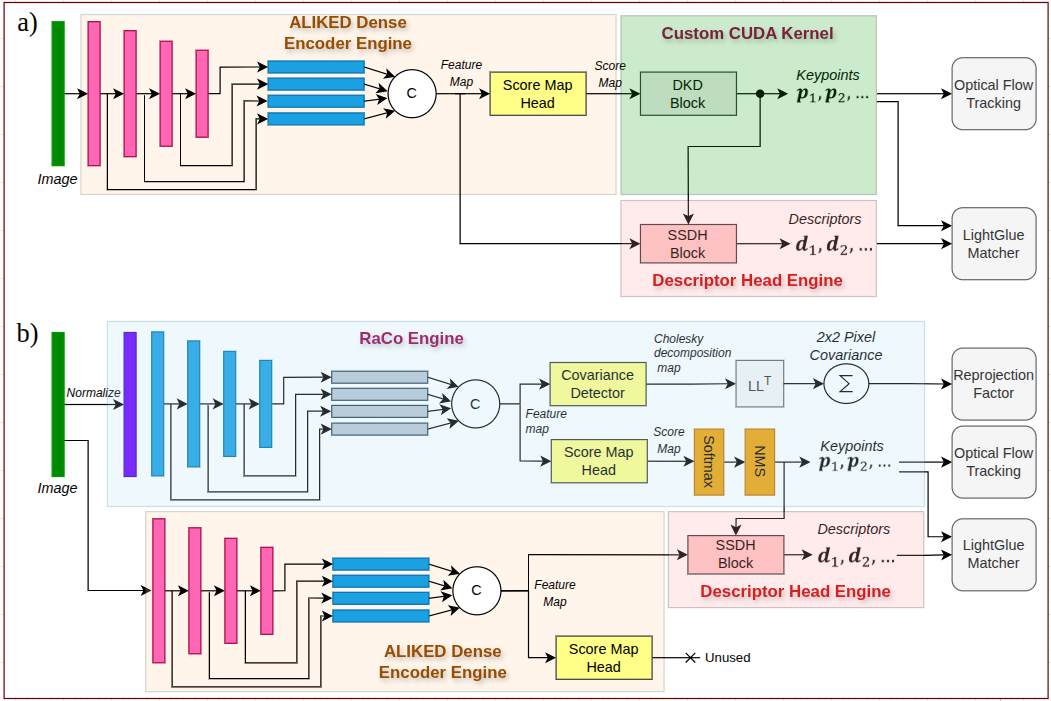}
  \caption{a) ALIKED model serialized into two independent engines, with custom CUDA Kernel replacing the DKD for keypoint selection. b) RaCo reused the ALIKED decomposed engine for generating descriptors over its keypoints.}
  \label{fig:decomposition}
\end{figure*}

We provide a brief overview of each front-end model currently assessed in our framework:
\begin{itemize}
  \item \textbf{ALIKED}~\cite{aliked}: A lightweight network processing 3-channel RGB input that utilizes a Sparse Deformable Descriptor Head (SDDH) to extract 64- or 128-dimensional descriptors with strong geometric invariance. It generates a score map processed by a Differentiable Keypoint Detection (DKD) module to yield accurate and differentiable subpixel keypoints.
  \item \textbf{RaCo}~\cite{raco}: A dedicated keypoint detector for 3-channel RGB input. It uniquely integrates a multi-scale covariance estimator to quantify spatial uncertainty and a differentiable ranking module to globally prioritize the most repeatable keypoints.
  \item \textbf{SuperPoint}~\cite{SuperPoint}: A self-supervised Cnvolutional Neural Network (CNN) with dual decoders that processes 1-channel grayscale images. It outputs a 256-dimensional descriptor and extracts discrete keypoints via Non-Maximum Suppression (NMS) from a dense probability map.
  \item \textbf{XFeat}~\cite{xfeat}: An ultra-lightweight shallow CNN optimized that processes 1-channel grayscale input to output a compact 64-dimensional descriptor. Keypoints are detected via a dedicated 1x1 convolution fast branch and are uniquely filtered using an additional reliability heatmap that estimates confidence.
\end{itemize}

The core of the framework rests on a single notion. For any learned visual front-end $\Phi_\theta$ with frozen weights $\theta$, the model maps an image to a structured feature set:
\begin{equation}
  \Phi_\theta(\mathbf{I}) \;=\; \bigl\{(\mathbf{p}_i,\; \mathbf{d}_i,\; s_i)\bigr\}_{i=1}^{N},
  \label{eq:frontend_abstraction}
\end{equation}
where image $\mathbf{I}$ yields $N$ keypoints with pixel coordinates $\mathbf{p}_i \in \mathbb{R}^2$, $\ell_2$-normalized descriptors $\mathbf{d}_i \in \mathbb{R}^D$ (dimension, $D \in \{64,128,256\}$ by extractor), and detection scores $s_i \in [0,1]$. The scores are processed internally for keypoints selection. We write per-keypoint quantities in lower-case bold symbols and their stacked counterparts in upper-case boldface ($\mathbf{P}\in\mathbb{R}^{N\times2}$, $\mathbf{D}\in\mathbb{R}^{N\times D}$, $\mathbf{S}\in\mathbb{R}^{N}$). This abstraction lets any conforming extractor be swapped without changing downstream modules.

RaCo detects keypoints but emits no descriptors, and its public checkpoint ships a LightGlue matcher trained on RaCo keypoints with ALIKED descriptors. We therefore decompose ALIKED in our exporter to supply descriptors over RaCo's keypoints, as illustrated in Fig.~\ref{fig:decomposition}. The feature set in Eq.~\eqref{eq:frontend_abstraction} for RaCo is then assembled by conditioning ALIKED's descriptor head on RaCo keypoints:
\begin{equation}
\begin{aligned}
  (\mathbf{P}_{\mathrm{Ra}}, \mathbf{S}_{\mathrm{Ra}})
    &= \mathcal{D}_{\theta_{\mathrm{Ra}}}(\mathbf{I}),\\
  \mathbf{F}_{\mathrm{AL}}
    &= \mathcal{E}_{\theta_{\mathrm{AL}}}(\mathbf{I})
       \in \mathbb{R}^{H \times W \times 128},\\
  \mathbf{D}_{\mathrm{Ra}\leftarrow\mathrm{AL}}
    &= \mathcal{H}_{\theta_{\mathrm{AL}}}
       (\mathbf{F}_{\mathrm{AL}}, \mathbf{P}_{\mathrm{Ra}}),\\
  \Phi_{\mathrm{Ra}\leftarrow\mathrm{AL}}(\mathbf{I})
    &= \bigl\{(\mathbf{p}_i^{\mathrm{Ra}},
        \mathbf{d}_i^{\mathrm{Ra}\leftarrow\mathrm{AL}},
        s_i^{\mathrm{Ra}})\bigr\}_{i=1}^{N_{\mathrm{Ra}}}.
\end{aligned}
\label{eq:raco_aliked_composite}
\end{equation}
Here the frozen RaCo detector $\mathcal{D}_{\theta_{\mathrm{Ra}}}$ returns stacked keypoints $\mathbf{P}_{\mathrm{Ra}} \in \mathbb{R}^{N_{\mathrm{Ra}} \times 2}$ and scores $\mathbf{S}_{\mathrm{Ra}} \in \mathbb{R}^{N_{\mathrm{Ra}}}$. The frozen ALIKED dense encoder $\mathcal{E}_{\theta_{\mathrm{AL}}}$ produces its dense feature map $\mathbf{F}_{\mathrm{AL}}$, and the SDDH descriptor head $\mathcal{H}_{\theta_{\mathrm{AL}}}$ takes this feature map together with the RaCo keypoints $\mathbf{P}_{\mathrm{Ra}}$ to extract one descriptor per keypoint, $\mathbf{D}_{\mathrm{Ra}\leftarrow\mathrm{AL}}$. The subscript $\mathrm{Ra}\!\leftarrow\!\mathrm{AL}$ thus denotes RaCo locations and scores paired with ALIKED descriptors, while the output count $N_{\mathrm{Ra}}$ is set by RaCo alone.

For LK optical-flow tracking, which needs keypoints only, we skip the ALIKED descriptor head for both RaCo and ALIKED pathway. SuperPoint and XFeat are deployed unmodified.

Two implementation details warrant attention. First, we replace ALIKED's DKD stage with a custom CUDA kernel. Since the DKD emits dynamically shaped output, isolating it lets the dense encoder and descriptor head export as static-shape TensorRT engines. The kernel fuses the original PyTorch NMS, sorting, and patch extraction into three single-stream stages, so refinement cost scales with the keypoint count $N$ rather than the image area $H\times W$ and admits GPU-friendly loop unrolling.

Second, RaCo's covariance estimator and ranking head are inactive in our initial export. We re-enable the covariance head to predict a $2\times2$ spatial-uncertainty matrix per keypoint. Following inspiration from MAC-VO~\cite{macvo}, this covariance forms an anisotropic square-root information matrix for the reprojection factor in place of the VINS-Fusion isotropic weight, downweighting poorly localized keypoints while preserving directional uncertainty. We evaluate it as a controlled ablation in Sec.~\ref{sec:raco_cov_ablation} and provide the formulation in Appendix~\ref{app:cov}.

\subsection{Input pre-processing kernel}
\label{subsec:preproc}

To minimize host--device transfer overhead, a fused CUDA pre-processing kernel handles letterbox padding, color-format conversion, intensity normalization, and NCHW layout packing in a single pass, with stereo pairs processed as one batch. Because the TensorRT engines are exported with static input shapes, frames are padded to the nearest valid resolution grid (multiples of $8$ for SuperPoint and XFeat, $32$ for ALIKED and RaCo). Keypoints extracted in this padded space are shifted back to native coordinates by subtracting the offset. Monochrome inputs are replicated to three channels for ALIKED and RaCo, while color inputs are reduced to a single luminance channel for SuperPoint and XFeat.

\subsection{Deep Learning Feature Matching}
\label{subsec:tracking}

In place of optical flow, temporal tracking make use of frame to frame LightGlue matching~\cite{lightglue}. The current features $\mathcal{F}_t$ are matched against a cache of the previous frame's \emph{tracked} features $\mathcal{F}_{t-1}$, each tagged with its track
identity. Both sets stay resident on the GPU, so only the compact match indices return to the host. In stereo mode, the
temporal pair $(\mathcal{F}_{t-1}^L,\mathcal{F}_t^L)$ and the stereo pair $(\mathcal{F}_t^L,\mathcal{F}_t^R)$ are identically shaped and independent, and resolve in one batch-2 LightGlue launch rather than two sequential inferences, mirroring the batch-2 extractor whose single pass already yields $(\mathcal{F}_t^L,\mathcal{F}_t^R)$. Temporal matches propagate track identities across the sliding window whereas stereo matches anchor per-frame depth.

Each feature set follows the abstraction in Eq.~\eqref{eq:frontend_abstraction},
$\mathcal{F}_t^{c}=\Phi_\theta(\mathbf{I}_t^{c})$ for camera $c\in\{L,R\}$. The matcher, LightGlue in this case, is then a map
between two such sets,
\begin{equation}
  \mathrm{LG}(\mathcal{F}^{L},\mathcal{F}^{R}) =
  \bigl\{\,(i,j)\;:\;w_{ij}\ge\beta\,\bigr\},
  \label{eq:lg_map}
\end{equation}
i.e., each keypoint of $\mathcal{F}^L$ matches at most one keypoint of $\mathcal{F}^R$, kept when the confidence $w_{ij}\in[0,1]$ exceeds a threshold $\beta$. As the matcher engine is exported with a static shape, each set is zero-padded to the extractor's keypoint cap $N_{\max}$. This yields the temporal set $\mathcal{M}_T=\mathrm{LG}(\mathcal{F}_{t-1}^{L},\mathcal{F}_t^{L})$ and the stereo set $\mathcal{M}_S=\mathrm{LG}(\mathcal{F}_t^{L},\mathcal{F}_t^{R})$.

During initialization, the track cache is empty. Therefore, all initial detections instantiate new feature tracks. In subsequent frames, detections successfully associated via $\mathcal{M}_T$ retain their cached identities, while unmatched detections spawn new tracks. The cache is subsequently updated utilizing the current left-frame features. Temporal correspondences then undergo geometric verification via MAGSAC++~\cite{magsac}. Specifically, matched keypoints are transformed to the normalized image plane for robust essential matrix estimation, and correspondences exhibiting a point-to-epipolar-line error exceeding a strict 1-pixel threshold are rejected.

Stereo correspondences are instead validated directly through calibrated geometry. Matches with negligible disparity are discarded. The remaining pairs are triangulated using the known or online estimated stereo baseline and are retained exclusively if they yield a positive depth within a valid range and exhibit a reprojection error of no more than 1.5-pixel across both views. Following geometric verification, the refined observations are forwarded to the backend. Finally, a sliding-window estimator performs a joint optimization over the reprojection residuals and IMU preintegration terms.

\subsection{Loop Closure}

For long-term drift correction, we add a front-end-agnostic loop-closure module. Each keyframe is encoded by a frozen DINOv2 ViT-S/14 network, whose dense patch tokens are aggregated by a universal VLAD codebook with \(K=32\) centers into an \(\ell_2\)-normalized global descriptor \(\mathbf{g}\in\mathbb{R}^{K\cdot384}\), following AnyLoc~\cite{AnyLoc}.

Loop candidates are retrieved using cosine similarity between the current descriptor \(\mathbf{g}_t\) and a past descriptor \(\mathbf{g}_k\):
\begin{equation}
  \mathrm{sim}(\mathbf{g}_t,\mathbf{g}_k)
  =
  \frac{\mathbf{g}_t^{\top}\mathbf{g}_k}
  {\|\mathbf{g}_t\|_2\|\mathbf{g}_k\|_2}
  =
  \mathbf{g}_t^{\top}\mathbf{g}_k ,
  \label{eq:retrieval}
\end{equation}
where the final equality follows from unit-\(\ell_2\) normalization. Loop keyframes are inserted according to travelled distance, and a fixed window of the most recent keyframes is excluded from retrieval to avoid self-matches. Candidates with \(\mathrm{sim}(\mathbf{g}_t,\mathbf{g}_k)>\tau_{\mathrm{sim}}\) are ranked by similarity, and the first geometrically verified match is added as a loop constraint to the inherited pose-graph back-end.

Geometric verification reuses the active local descriptors of the selected front-end through cosine nearest-neighbour matching with a Lowe ratio test, followed by 2D--2D fundamental-matrix RANSAC and 3D--2D PnP RANSAC. Retrieval is implemented as a brute-force scan over the keyframe database, with \(\mathcal{O}(n)\) per-query cost. This is adequate for the trajectory lengths considered in our VI-SLAM evaluation, so approximate indexing libraries such as FAISS~\cite{faiss} are not used. Optical-flow variants, which do not maintain learned descriptors, fall back to the standard DBoW2 BRIEF loop-closure path.

\section{Experimental Setup}
\label{sec:setup}

\begin{table}[t]
\centering
\small
\renewcommand{\arraystretch}{1.2}
\caption{Evaluation Platform Specifications}
\label{tab:platforms}
\begin{tabular}{@{}lll@{}}
\toprule
 & \textbf{Workstation} & \textbf{Jetson AGX Orin} \\
\midrule
CPU    & Intel Core i7-12700H (20T)        & Arm Cortex-A78AE (12)  \\
GPU    & NVIDIA RTX 3080 Ti                & NVIDIA Ampere      \\
Memory & 64\,GB, 16\,GB (VRAM)             & 64\,GB (unified)   \\
\bottomrule
\end{tabular}
\end{table}

\begin{table*}[htbp]
\centering
\small 
\renewcommand{\arraystretch}{1.3} 
\caption{Evaluation Dataset Summary}
\label{tab:dataset_summary}

\begin{tabularx}{\textwidth}{@{} l l c >{\raggedright\arraybackslash}X >{\raggedright\arraybackslash}X >{\raggedright\arraybackslash}X @{}}
\toprule

\textbf{Dataset} & \textbf{Sequence} & \textbf{Total Length (m)} & \textbf{Sensors (Camera \& IMU Only)} & \textbf{Ground Truth} & \textbf{Scene description} \\

\midrule

EuRoC \cite{euroc} & 11 & 893.45 & 
\textbf{Cam:} 2$\times$ MT9V034 (Mono, Global Shutter, 752$\times$480 @ 20 Hz) \newline
\textbf{IMU:} ADIS16448 (200 Hz) & 
Vicon motion capture, Leica laser tracker & 
Indoor MAV flights (industrial hall \& motion capture room) with varying motion profile. \\

NTU-VIRAL \cite{ntu_viral} & 9 & 1845.24 & 
\textbf{Cam:} 2$\times$ uEye 1221 LE (Mono, Global Shutter, 752$\times$480 @ 10 Hz) \newline
\textbf{IMU:} VectorNav VN100 (385 Hz) & 
Leica Nova total station (laser tracking prism) & 
Indoor and outdoor UAV flights on a university campus with structural variations. \\

Botanic Garden$^\dagger$ \cite{botanic_garden} & 7 & 3504.73 &
\textbf{Cam:} Teledyne DALSA M1930/C1930 (Mono \& RGB, Global Shutter, 1920$\times$1200 @ 40 Hz) \newline
\textbf{IMU:} Xsens Mti-680G (400 Hz) & 
Leica RTC360 3D scanner map \& map-based localization & 
Unstructured natural environments (woods, trails, riversides, dense vegetation) using a wheeled robot. \\

SubT-MRS$^\ddagger$ \cite{subt_mrs} & 5 & 1609.70 & 
\textbf{Cam:} Leopard Imaging LI-XAVIER Fisheye (RGB, 686$\times$816 @ 24 Hz) \newline
\textbf{IMU:} Epson M-G365 (200 Hz) & 
Centimeter-level multi-modal mapped trajectories (FARO 3D scanner) & 
Subterranean (caves, tunnels) and urban environments with extreme multi-degraded conditions (smoke, dust, darkness). Heterogeneous robots. \\

\bottomrule
\multicolumn{6}{@{}l}{\footnotesize 1) $^\dagger$ Subset of the dataset utilizing the provided downsampled rosbag image resolutions (960x600@10Hz).} \\
\multicolumn{6}{@{}l}{\footnotesize 2) $^\ddagger$ Evaluation performed on visual track challenge subsets with provided rosbag image resolution (640x480@24Hz).} \\
\end{tabularx}
\end{table*}

Experiments are conducted on the two hardware platforms summarized in Table~\ref{tab:platforms}. Both platforms run CUDA~12.6 and TensorRT~10.3. Since TensorRT engines are platform-dependent, all engines are serialized separately for each target and are not portable across devices. We evaluate DL-VINS-Factory on 4 public benchmarks spanning diverse environments, motion profiles, and sensor configurations, as summarized in Table~\ref{tab:dataset_summary}. 

All learned models are deployed using their original pre-trained weights, without retraining or fine-tuning on any evaluation dataset. Likewise, the VLAD codebook used for loop-closure retrieval is not constructed from the evaluation datasets. SuperPoint weights are taken from the original MagicLeap release, although a TensorFlow reimplementation with a more permissive license variant exists, it is not used in this work. For ALIKED, we use the N16 variant, which provides a representative balance between accuracy and throughput compared with the rotation-augmented N16rot, the lightweight T16, and the larger N32 models. All learned front-end models are exported to ONNX, simplified using \texttt{onnxsim\footnote{https://github.com/onnxsim/onnxsim}}, and serialized into platform-specific TensorRT engines with FP16 precision and capped 256 keypoint output. During runtime, each engine is warmed up using dummy inputs before processing rosbag data.

All experiments utilized the default calibration profiles provided by each dataset which consist of camera intrinsics, IMU noise parameters, and IMU--camera extrinsics. We include Shi-Tomasi Corner~\cite{gfft} with LK optical-flow tracking \cite{lkoptflow}, and SIFT~\cite{sift} with either LK optical flow tracking or LightGlue matching as controlled baselines. Together with the 4 learned extractors, this yields 11 front-end configurations, which we evaluate across 28 sequences spanning the four datasets. Every sequence is run in both mono- and stereo-inertial setups, except SubT-MRS, whose rosbags record only a single camera. In addition, each Botanic Garden sequence is evaluated separately under both its RGB and grayscale cameras. VINS-Fusion~\cite{vins_fusion} and OKVIS2~\cite{okvis2} are also evaluated as system-level baselines.

Localization accuracy is reported using the root-mean-square error (RMSE) of the Absolute Trajectory Error (ATE), computed with the \texttt{evo}\footnote{https://github.com/michaelgrupp/evo} toolkit. On the Jetson AGX Orin, runtime performance is monitored with the built-in \texttt{tegrastats} utility.

\section{Results and Analysis}
\label{sec:results}
For the remainder of the manuscript, we denote Good Features to Track as GFTT, ALIKED as AL, RaCo as Ra, SuperPoint as SP, XFeat as XF, Lucas-Kanade optical flow as LK, and LightGlue as LG. The notation Ra+LG denotes the RaCo extractor paired with LightGlue, and likewise for further combinations.

\subsection{Accuracy Analysis}
\label{sec:accuracy analysis}

\begin{table}[t]\centering\scriptsize\setlength{\tabcolsep}{3pt}\renewcommand{\arraystretch}{1.2}
\begin{threeparttable}
\caption{Mean absolute trajectory error (ATE RMSE [m]) per dataset, averaged over sequences. Each cell represents odometry / loop-closed trajectory. Measured on x86\_64 platform.Per-sequence results in Appendix~\ref{app:ate}.}
\label{tab:ate_summary}
\begin{tabular}{l *{4}{c}}\toprule
Method & EuRoC & NTU-VIRAL & Botanic-Gray & Botanic-RGB \\ \midrule
\multicolumn{5}{@{}l}{\textbf{Mono-Inertial}}\\[1pt]
VINS-Fusion & 0.158$^{\dagger}$/0.141$^{\dagger}$ & 0.933$^{\dagger}$/0.604$^{\dagger}$ & 1.295$^{\dagger}$/1.267$^{\dagger}$ & 1.299/1.305${^*}$ \\
OKVIS2 & 0.162/\sbest{0.073} & \ding{55} & 6.463/6.510${^*}$ & 3.612/3.764${^*}$ \\
\cmidrule(lr){1-5}
GFTT (LK) & \sbest{0.154}/0.084 & 1.260/0.709 & 1.377/1.194 & 1.200/1.095 \\
SIFT (LK) & 0.170/0.089 & 1.476/0.849 & 1.704/1.511 & 1.607/1.540 \\
SIFT (LG) & 0.563$^{\dagger}$/0.141$^{\dagger}$ & 1.295$^{\dagger}$/0.856$^{\dagger}$ & 4.728$^{\dagger}$/4.691$^{\dagger}$ & 1.994$^{\dagger}$/1.984$^{\dagger}$ \\
AL (LK) & 0.176/0.159 & 1.400/1.288 & 1.344/1.257 & 1.200/1.076 \\
AL (LG) & \best{0.146}/\best{0.070} & 0.978/\sbest{0.544} & 2.051/1.812 & 1.380/1.167 \\
Ra (LK) & 0.166/0.080 & 1.110/0.621 & \sbest{1.109}/\sbest{0.888} & \best{0.767}/\best{0.681} \\
Ra (LG) & 0.168/0.087 & \best{0.877}/\best{0.513} & 1.526/1.257 & 1.095/0.975 \\
SP (LK) & 0.162/0.078 & 1.172/0.680 & \best{1.085}/\best{0.850} & 0.969/\sbest{0.838} \\
SP (LG) & 0.176/0.098 & \sbest{0.973}/0.624 & 4.343$^{\dagger}$/4.020$^{\dagger}$ & 1.577$^{\dagger}$/1.348$^{\dagger}$ \\
XF (LK) & 0.157/0.093 & 1.063$^{\dagger}$/0.720$^{\dagger}$ & 1.124$^{\dagger}$/0.966$^{\dagger}$ & \sbest{0.942}/0.883 \\
XF (LG) & \ding{55} & \ding{55} & \ding{55} & \ding{55} \\
\midrule
\multicolumn{5}{@{}l}{\textbf{Stereo-Inertial}}\\[1pt]
VINS-Fusion & 0.222/0.185 & 0.649/0.486 & 2.615/2.626${^*}$ & \sbest{2.769}/2.716 \\
OKVIS2 & \best{0.060}/\best{0.030} & 1.452/1.034 & 2.338/2.404${^*}$ & \best{1.674}/\best{1.644} \\
\cmidrule(lr){1-5}
GFTT (LK) & \sbest{0.122}/0.068 & 0.711/\sbest{0.402} & 2.747/2.329 & 3.365/3.227 \\
AL (LK) & 0.140/0.123 & 0.680/0.647 & 2.917/2.893 & 3.129/2.988 \\
AL (LG) & 0.127/\sbest{0.063} & \best{0.552}/\best{0.354} & \best{1.965}/\best{1.839} & 3.184/\sbest{2.681} \\
Ra (LK) & 0.144/0.111 & 0.775/0.717 & 3.267$^{\dagger}$/3.233$^{\dagger}$ & 3.280/3.123 \\
Ra (LG) & 0.181/0.082 & 0.902/0.619 & \sbest{2.088}/\sbest{2.004} & 3.378/3.024 \\
SP (LK) & 0.129/0.104 & 0.708/0.647 & 3.293/3.238 & 3.077/2.950 \\
SP (LG) & 0.125/0.078 & \sbest{0.608}/0.402 & 2.818$^{\dagger}$/2.555$^{\dagger}$ & 3.869/3.097 \\
XF (LK) & 0.149/0.113 & 0.734/0.666 & 3.424/3.418 & 2.951/2.778 \\
XF (LG) & 0.377/0.340 & 2.315/2.315 & \ding{55} & 1.924$^{\dagger}$/1.897$^{\dagger}$ \\
\bottomrule
\end{tabular}
\atenotes
\end{threeparttable}\end{table}

\begin{figure*}[t]
  \includegraphics[width=1.0\textwidth]{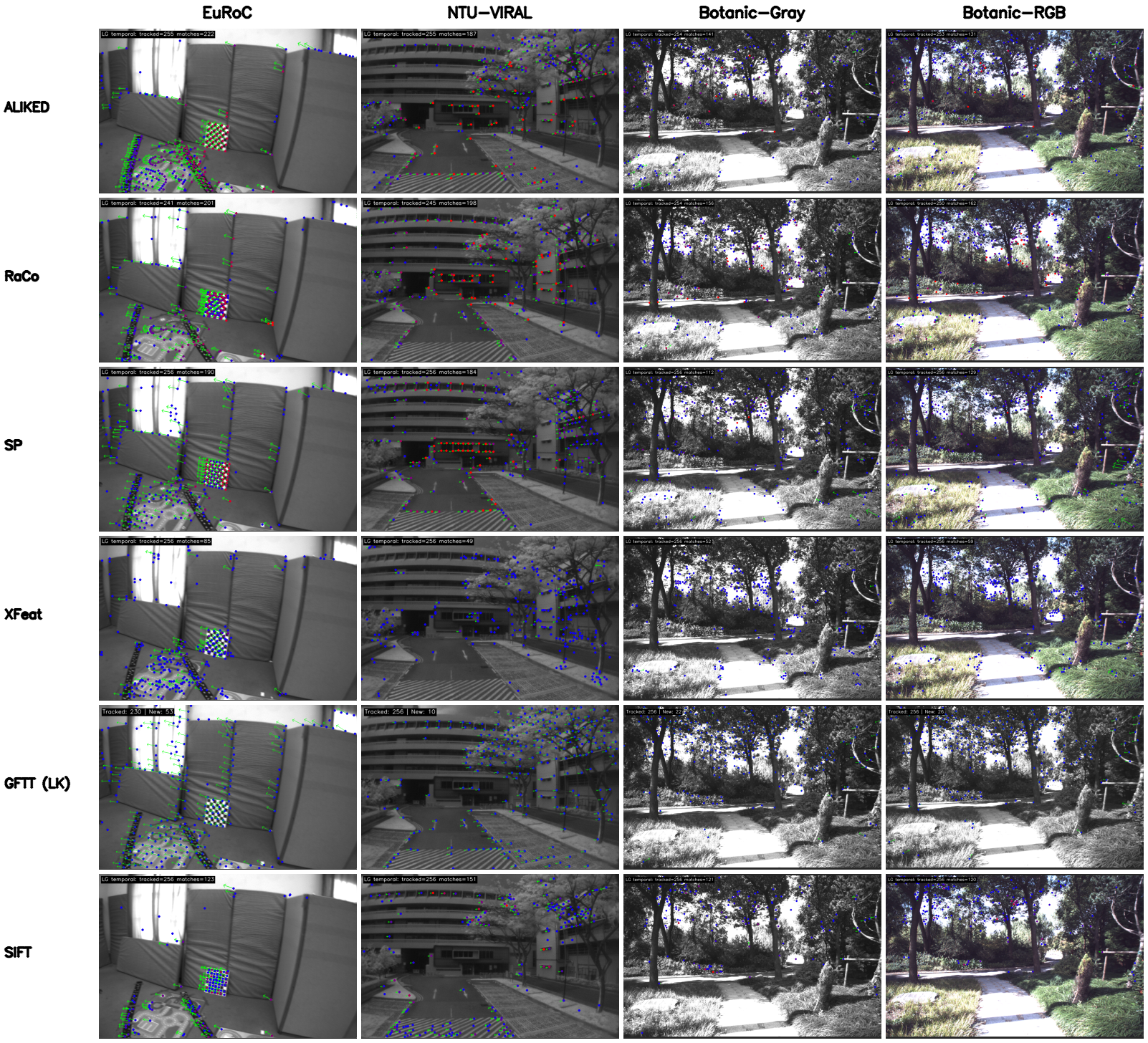}
  \caption{Qualitative comparison of feature distribution and temporal tracking across extractors and datasets. Blue points show tracked current-frame features, red points indicates LightGlue correspondences from previous frame, and green vectors denote frame-to-frame 2D pixel displacement.}
  \label{fig:track_grid}
\end{figure*}

This section mainly addresses the visual-inertial \emph{odometry} accuracy. The impact and failure modes of loop-closure additions are deferred to Sec.~\ref{sec:loopclosure}. Each cell of Table~\ref{tab:ate_summary} reports odometry / loop-closed ATE (RMSE), and unless noted, we quote the odometry value in this section. Per-sequence results are given in Appendix~\ref{app:ate}. Based on the evaluation outcome, we observed that \emph{no single front-end achieves superiority in tested environments}. Rather, the most effective configuration is governed by the underlying scene structure, the extent of viewpoint variation, and specific ego-motion characteristics, independent of the feature extractor's inherent representational capacity.

EuRoC remains the gold standard for benchmarking visual odometry, though it is largely solved under its well-lit, structured conditions. OKVIS2 yields the lowest stereo odometry ATE (0.060 m) due to its 700-keypoint BRISK upper limit and bounded factor-graph formulation. Among the learned monocular setups capped at 256 features, AL+LG is optimal (0.146 m), yet all configurations cluster within 0.146--0.176 m. Crucially, the classical GFTT+LK baseline attains the second-lowest odometry ATE in \emph{both} the monocular (0.154 m) and stereo (0.122 m) configurations under the same feature cap, demonstrating that keypoint density and factor-graph formulation dictate performance on this dataset more than feature representation.

The NTU-VIRAL dataset poses distinct challenges from aerial and rotational dynamics, as well as distant scene structures. In accordance with the dataset evaluation configuration, we enabled online camera-IMU extrinsic estimation across all methods, including the OKVIS2 and VINS-Fusion baselines. Notably, this feature remains disabled for all other datasets in our study. For the DL-VINS front-ends, we further regularize this extrinsic estimation by applying a bounded prior factor to the stereo baseline. This prevents metric scale drift in distant scenes without rigidly enforcing the dataset's initial calibration. Additionally, for OKVIS2 in outdoor environments, including the Botanic Garden datasets, we activated its CNN-based segmentation to mask the sky and dynamic pedestrians. Under these conditions, we found that Ra+LG achieves the lowest monocular ATE (0.877 m), while AL+LG attains the lowest stereo ATE (0.552 m). In contrast, all monocular optical flow variants yield errors between 1.06 m and 1.40 m. While OKVIS2 fails across all monocular configurations, it successfully tracks in stereo (1.452 m), indicating that its segmentation and online extrinsic handling only provide a net benefit when stereo vision supplies observable depth. Consequently, given the relatively low camera frame rate (10 Hz), the learned matching proves more effective in handling the large inter-frame displacements, conditions under which optical flow tracking may tend to struggle.

The Botanic Garden dataset sharply reverses the stereo and LightGlue matching narrative observed elsewhere. Monocular tracking peaks with optical flow variants SP+LK (grayscale: 1.085 m) and Ra+LK (RGB: 0.767 m), while stereo degrades performance by a factor of two across most learned front-ends. OKVIS2 is the exception, achieving the best RGB stereo ATE (1.644 m) despite failing in monocular grayscale (6.463 m). This degradation is primarily caused by the low stereo match yield in open vegetation. The stereo pipeline recovers only 125--132 valid correspondences per frame, approximately half of the 256-keypoint cap and substantially fewer than the roughly 200 correspondences observed on EuRoC. Moreover, the surviving matches exhibit low effective disparity. As a result, the stereo and cross-camera factors conflict with the monocular temporal factors during optimization. For a ground robot following near-straight trajectories, these weak stereo constraints pollute the estimator rather than improve it.

Furthermore, minor translational offsets between the dataset's separate monochrome and color cameras heavily influence performance. For instance, in sequence 1005\_07, monocular GFTT yields 1.891 m on grayscale versus 1.444 m on color. Although the detectors compute keypoint responses from local image neighborhoods, the displayed tracks are also affected by global selection steps such as thresholding, NMS, top-$N$ retention, and descriptor matching. Hence, the Botanic--Gray and Botanic--RGB views can exhibit different tracked-keypoint distributions despite the small camera mounting position difference. Their different spectral responses, exposure and image-processing pipelines alter local gradients and texture contrast, while vegetation-induced parallax and partial occlusion further affect which features are detected, matched, and retained as tracks. A full qualitative comparison of the keypoints extracted on similar frames from different datasets is presented in Fig. \ref{fig:track_grid}.

LightGlue matching is further penalized by weak left-right correspondences, causing under-constrained windows and divergence, especially for SP+LG on the 1005\_07 sequence. Monocular SP+LG also suffers long 38-second initialization delays on the low-parallax sequence 1018\_00, compared to under 5 seconds for other methods, because they fail to maintain the long-track counts necessary to cross the initialization parallax gate. On the same sequence, Ra+LK stereo diverges completely despite Ra+LK mono succeeding at 0.544 m and Ra+LG stereo tracking at 0.596 m. Feature-tracker diagnostics show that RaCo yields relatively lower detections at around 38 per frame in the featureless vegetation corridor, against 180--250 for all other detectors in the early segment. In stereo mode, only features that pass cross-camera optical flow verification enter the temporal pool, and with so few candidates, it is unable to initialize the estimator on this low-parallax sequence. In mono mode, this gate is absent, so the pool self-sustains via temporal carry-over at a 95\% forward rate and crosses the initialization parallax gate. Finally, stereo-inertial performance on sequence 1008\_03 reveals an anomaly where RGB errors double those of grayscale, likely due to corrupted RGB recordings operating at roughly 4 Hz compared to the monochrome's 10 Hz.

XF+LG produces no valid trajectory in any monocular configuration across the four datasets and also fails in the stereo-inertial configuration on Botanic Garden. Frame-level diagnostics indicate that new-track instantiation fails to establish sufficient matches in subsequent frames, causing the tracked feature set to decay and preventing initialization of the sliding-window backend. In stereo-inertial mode, XF+LG can initialize through the stereo anchor, but its accuracy remains poor on the datasets where it does not fail outright, with stereo ATEs of 0.377 m on EuRoC and 2.315 m on NTU-VIRAL. This failure mode is specific to the XF+LG pathway, likely reflecting limitations of its “LighterGlue'' matcher, which uses a smaller backbone than the other LightGlue-based front-ends. The XFeat extractor itself remains viable when paired with optical-flow tracking, indicating that the degradation originates primarily from the XFeat-specific learned matching branch. In contrast, SIFT is omitted from the stereo pipeline because successive extraction on both views of each stereo frame does not satisfy the real-time constraint.

Two notable outliers share a Ra+LG stereo failure signature. On EuRoC V203, Ra+LG stereo obtains 0.845 m (3.6x the column median of 0.233 m), while the paired Ra+LK stereo achieves 0.163 m. A second instance occurs on NTU-VIRAL SBS02, where RaCo+LG stereo yields 3.357 m (4.6x the column median of 0.724 m) against Ra+LK stereo at 0.724 m. In both cases, LightGlue stereo matching over RaCo keypoints, when combined with borrowed ALIKED descriptors, appears to produce a higher fraction of geometrically inconsistent correspondences in certain high-motion sequences. The resulting loss of valid triangulations erodes metric scale estimation during steady-state tracking. Repeated trials on both sequences confirm this behavior is systematic rather than a marginal-stability artifact.

\begin{figure}[t]
  \includegraphics[width=0.5\textwidth]{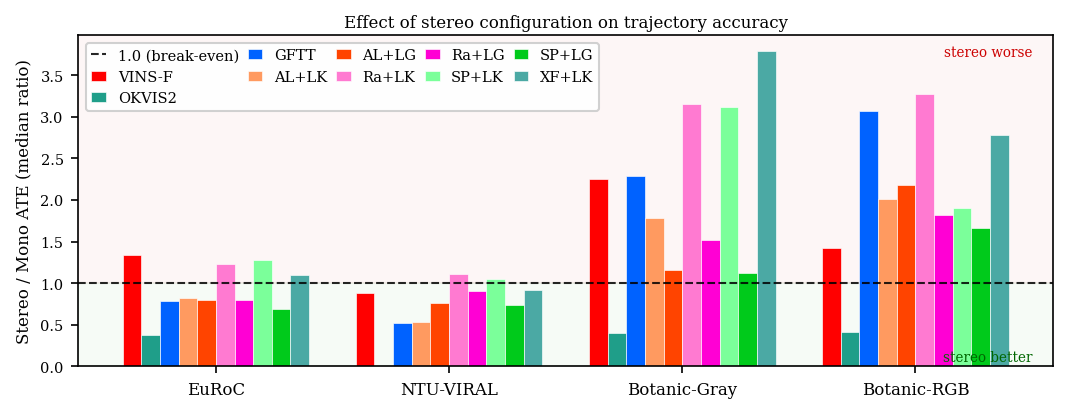}
  \caption{Median ATE ratio comparing stereo versus monocular accuracy across the datasets. The dashed line at 1.0 denotes equal performance. Values below 1.0 indicate better stereo performance, while values above 1.0 indicate otherwise.}
  \label{fig:stereo_vs_mono}
\end{figure}

\begin{figure}[t]
  \includegraphics[width=0.5\textwidth]{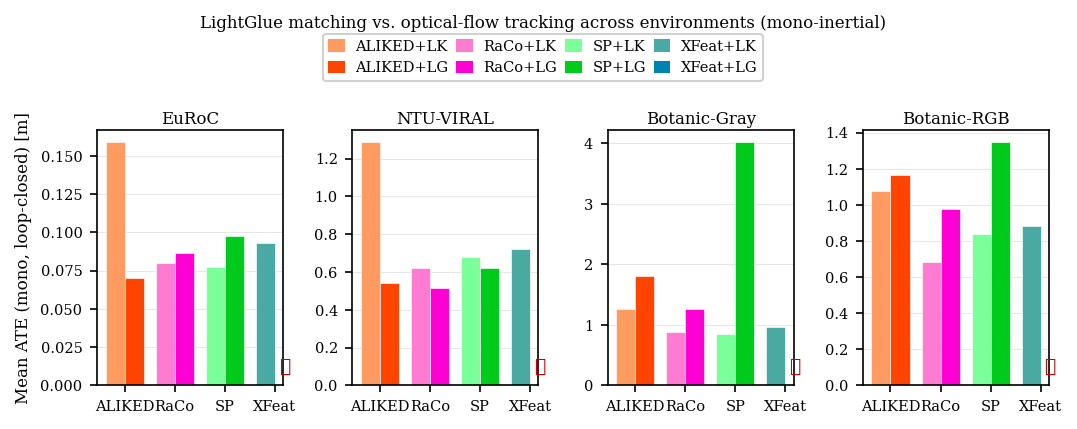}
  \caption{Mean ATE comparing Lucas-Kanade optical flow (+LK) against LightGlue matching (+LG) in mono-inertial configurations across the datasets.}
  \label{fig:lg_vs_lk}
\end{figure}

These dataset-specific outcomes highlight two notable trends, as encapsulated in Figs.~\ref{fig:stereo_vs_mono} and~\ref{fig:lg_vs_lk}. First, stereo configurations improve accuracy wherever metric depth is reliably observed, such as in near-field structured environments (EuRoC) or distant but highly textured scenes (NTU-VIRAL). The observed performance gap in monocular setups is primarily an \emph{initialization} limitation rather than a tracking failure. Because monocular visual-inertial systems recover scale only under sufficient motion excitation, stereo setups maintain a distinct advantage until adequate parallax is achieved. Conversely, stereo provides no benefit and can actively degrade performance in environments where disparity is poorly constrained, such as the low-texture vegetation of the Botanic Garden datasets. Second, LightGlue increasingly outperforms optical flow as inter-frame viewpoint variations grow. This advantage is evident during aerial motions in the NTU-VIRAL dataset. In contrast, optical flow remains preferable for near-straight, low-parallax trajectories with minimal out-of-plane (z-axis) movement, such as the Botanic Garden trails.

\subsection{Robustness Analysis}
\label{sec:robustness analysis}

\begin{table}[t]\centering\scriptsize\setlength{\tabcolsep}{2.0pt}\renewcommand{\arraystretch}{1.2}
\begin{threeparttable}
\caption{SubT-MRS absolute trajectory error (ATE RMSE [m]) on visually degraded scenarios (Mono-Inertial).}
\label{tab:subt_ate}
\begin{tabular}{l *{5}{c}}\toprule
Method & Lowlight 1 & Lowlight 2 & Over Exposure & Flash Light & Smoke Room \\ \midrule
\begin{tabular}{@{}l@{}}LET-NET \cite{letnet}, \\ VINS-Mono *\end{tabular} & \best{1.019} & \best{1.126} & 1.911 & 2.341 & \best{3.757} \\\cmidrule(lr){1-6}
\cmidrule(lr){1-6}
GFTT (LK) & 3.486 & 2.611 & 0.860 & \sbest{0.173} & \ding{55} \\
AL (LK) & 7.429 & 2.742 & \sbest{0.641} & 0.350 & \ding{55} \\
AL (LG) & 15.963 & 2.465 & 1.181 & 0.230 & \ding{55} \\
Ra (LK) & 16.403 & 2.013 & \ding{55} & 0.195 & \ding{55} \\
Ra (LG) & 2.511 & \sbest{1.661} & \ding{55} & \best{0.150} & \ding{55} \\
SP (LK) & 2.640 & \ding{55} & 0.658 & 0.218 & \ding{55} \\
SP (LG) & \sbest{1.061} & 2.325 & \ding{55} & \ding{55} & \ding{55} \\
XF (LK) & 1.497 & 2.550 & \best{0.638} & 0.224 & \ding{55} \\
XF (LG) & \ding{55} & \ding{55} & \ding{55} & \ding{55} & \ding{55} \\
\bottomrule
\end{tabular}
\begin{tablenotes}[flushleft]\footnotesize
  \item 1) Each cell reports the best between odometry and with loop-closed.
  \item 2) * Results are quoted directly from \cite{subt_mrs}.
\end{tablenotes}
\end{threeparttable}\end{table}

SubT-MRS provides the most adversarial conditions in this study, combining persistent low illumination, transient over-exposure and flash, and visual occlusion by smoke, all captured with a single monocular-inertial fisheye camera. In this scenario, the primary robustness indicator is sequence completion. Table~\ref{tab:subt_ate} reports, for each method, the better result between the odometry-only and loop-closed trajectories, and a \ding{55} denotes tracking loss in the pose estimation where more than 5\% deviation of the total run length. These sequences are evaluated with the nearest multiple of 32, which is a 160-keypoint upper limit as compared to the 256-keypoint cap used on the other datasets for fair comparison with the baseline LET-NET (a VINS-Mono derivative utilizing 150 keypoints that ranked first in pure online estimation). In this setup, no combination manages to traverse all five evaluated splits. Outdoor night is omitted due to missing ground truth, as well as the virtual sequence. Moreover, contrary to the common expectation that learned descriptors are uniformly more robust, several LightGlue variants are more brittle than lightweight optical-flow tracking, which can continue operating on the limited structure that remains visible.

\begin{figure}[t]
  \includegraphics[width=0.5\textwidth]{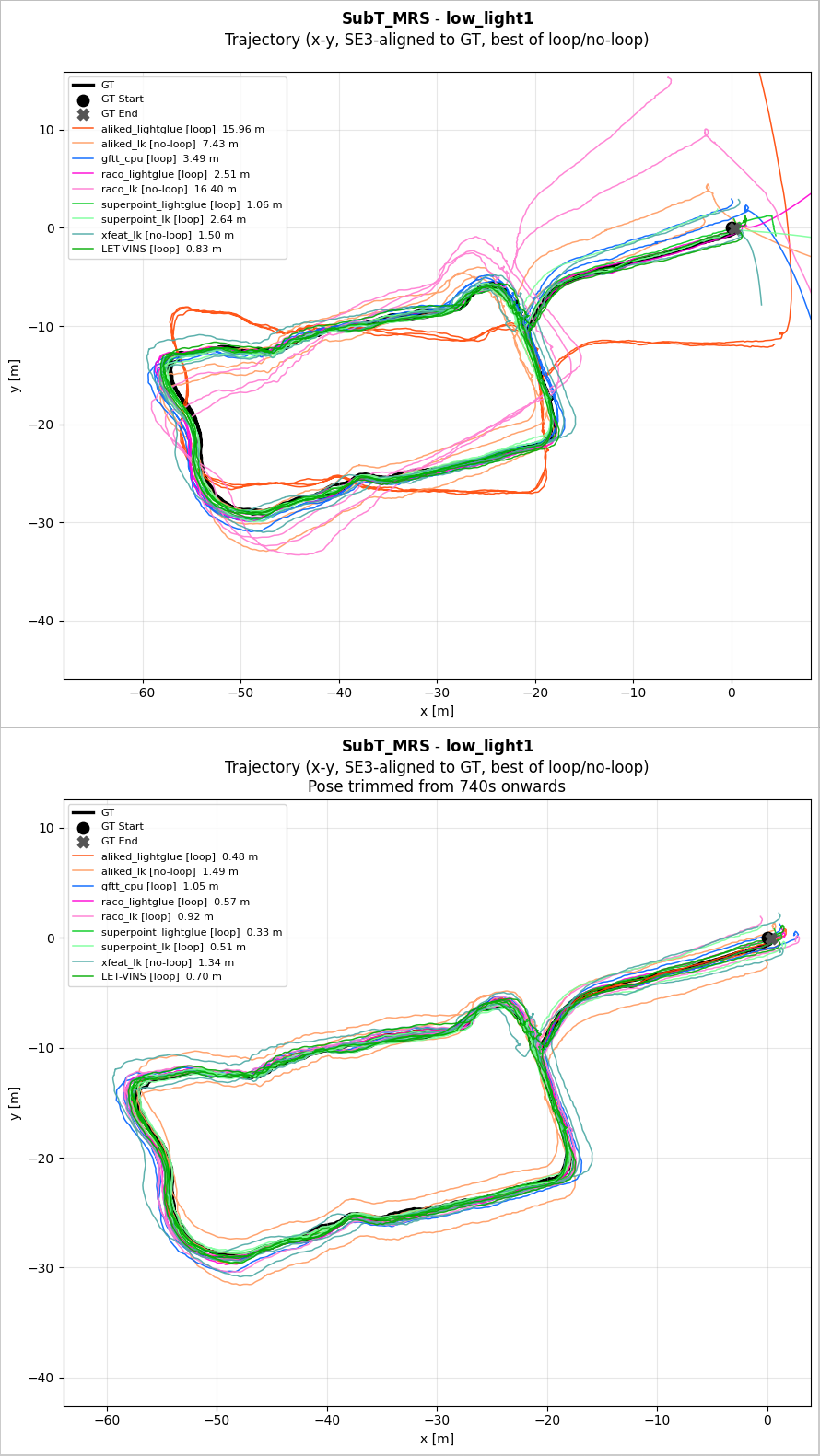}
  \caption{Trajectory evaluation on the $\mathbf{SubT\_MRS}$ \textbf{Lowlight 1} sequence, comparing various tracking methods against ground truth (GT) using SE3 alignment. The top plot shows the full length, where several methods experience significant divergence near the end of the run. The bottom plot displays the same trajectories with poses trimmed after 740 seconds, isolating the impact of the dynamically moving object while rig are stationary. The LET-NET, VINS-Mono baseline value shown here is from our own independent re-run.}
  \label{fig:lowlight1}
\end{figure}

Transient over-exposure and flashing light are the mildest challenges, learned extractors paired with optical flow remain accurate and outperform the baseline. XF+LK performed best with 0.638 m, while AL+LK is very close in second at 0.641 m in the over-exposure sequence, while Ra+LG is the best with 0.150 m on the flash light sequence. Conversely, the darkened cavern environments are substantially more demanding. The illumination-trained LET-NET baseline leads on the two Lowlight sequences, whereas SP+LG is the strongest learned configuration (1.061 m) in Lowlight 1 and Ra+LG is the strongest (1.661m) in Lowlight~2. There are configurations whose descriptors degrade on under-exposed frames, drifting severely (e.g.\ Ra+LK 16.4 m, AL+LG 16.0 m on Lowlight~1), although there is a distinct phenomenon worth highlighting. This catastrophic failure on Lowlight~1 is, however, not photometric but a sensitivity to dynamic objects in the camera viewpoint. During the final stretch, the sensor rig is held stationary, and the camera faces a seated person who, though not translating through the scene, makes repeated transient body movements that dominate the field of view. The resulting apparent motion drives the late-trajectory divergence for several methods. Trimming the trajectory after 740 s, from the total 807 s sequence, isolates this effect. The errors of Ra+LK and AL+LG collapse from 16.4 m and 16.0 m to 0.92 m and 0.49 m, as depicted in Fig.~\ref{fig:lowlight1}. For this sequence specifically, we additionally re-run the LET-NET baseline ourselves rather than quoting the paper as in Table~\ref{tab:subt_ate}, both to overlay its trajectory and to apply the same trimming. Our re-run measures 0.82\,m over the full sequence and 0.71\,m when trimmed. After trimming, several learned front-ends surpass this independently measured baseline, with SP+LG the best at 0.326\,m (AL+LG 0.49 m, SP+LK 0.51 m). This dynamic-object sensitivity is orthogonal to the illumination challenge.

Dense smoke poses the most severe challenge in the SubT-MRS dataset. Under the 160-feature cap, every learned front-end configuration loses tracking as airborne particulates obscure stable visual structure, forcing the estimator to rely primarily on IMU dead reckoning. This 160-feature limit is not intended to restrict front-end expressiveness, but rather to bound the number of features entering the sliding-window backend. This prevents marginalization and nonlinear optimization from becoming computationally prohibitive, thereby preserving real-time operation. Furthermore, this allocation size is particularly appropriate for SubT-MRS given its lower camera resolution compared to other benchmarks, alongside the application of a fisheye validity mask that culls features near the image periphery. Consequently, increasing the cap beyond this range is unlikely to yield proportional benefits, as extraneous detections would frequently be unavailable, poorly localized, or geometrically invalid. A detailed ablation study of this feature cap is provided in Sec.~\ref{sec:kp_ablation}.

Interestingly, increasing the cap to 256 features partially recovers performance for the optical flow variants; AL+LK achieves the best result (1.450 m), followed by SP+LK (2.792 m). This indicates that the tracking failures at the 160-feature limit are primarily driven by feature starvation under increasing visual occlusion, rather than an inherent limitation of the extractors. In contrast, no LightGlue-based configuration recovers reliably, even at the expanded 256-feature budget. AL+LG and Ra+LG diverge with errors of 6.5 m and 5.4 m, respectively, while SP+LG only marginally survives at the 5.0 m divergence threshold.

\begin{figure}[t]
  \includegraphics[width=0.5\textwidth]{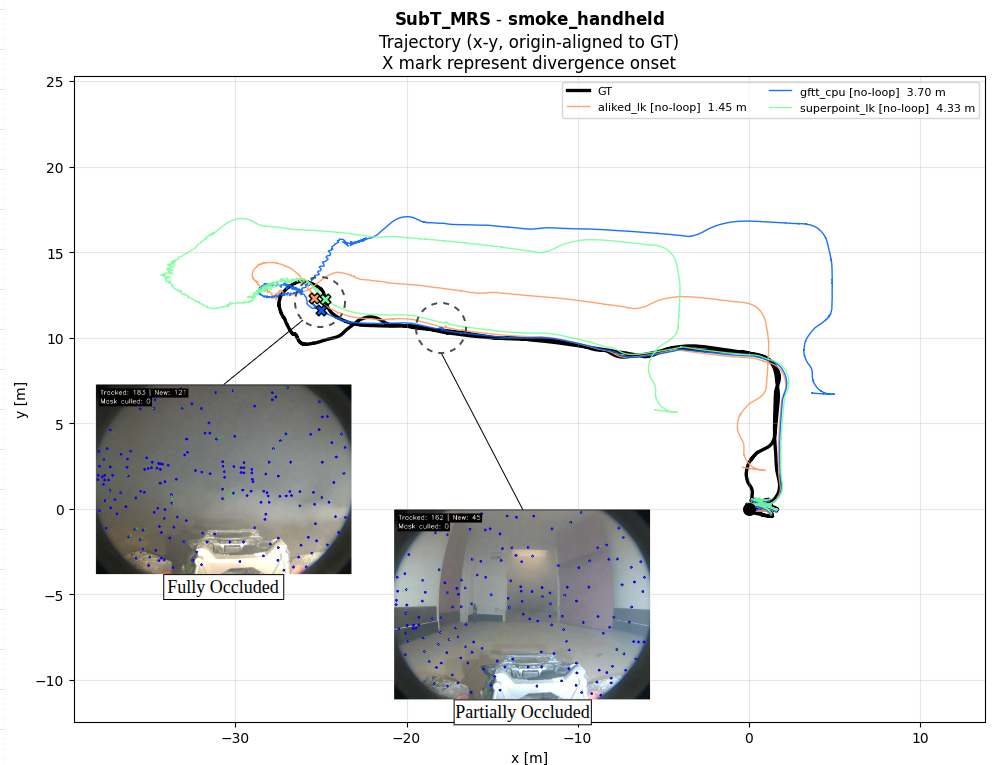}
  \caption{Trajectory evaluation on the $\mathbf{SubT\_MRS}$ \textbf{smoke room} sequence with origin aligned to ground truth (GT). Results are obtained with maximum extracted keypoints capped at 256. "X" markers indicate the onset of divergence, while inset images highlight the transition from partial to full visual occlusion.}
  \label{fig:smoke_room}
\end{figure}

Fig.~\ref{fig:smoke_room}, which presents trajectories origin-aligned to the ground truth, illustrates this failure mechanism. Most methods consistently track the initial, partially occluded segment but diverge at a common juncture corresponding to the transition into near-complete visual occlusion (highlighted by the image insets). At this stage, the near-total occlusion eliminates both stable corners and reliable learned features. Consequently, optical flow tracking degrades more gracefully by exploiting whatever sparse residual structure remains visible. LightGlue, however, relies on high-confidence descriptor correspondences and collapses entirely once the scene becomes severely obscured.

\subsection{Loop closure analysis}
\label{sec:loopclosure}

Loop closure followed a common retrieval, geometric verification and pose-graph correction pipeline. Keyframes are inserted by travelled distance, and the $50$ most recent keyframes are excluded to avoid trivial self-matches. Candidate loops are verified using their respective matching methods along with the fundamental matrix and PnP RANSAC, and accepted constraints require at least $25$ inliers before being accepted into the 4-DoF pose graph. For AnyLoc retrieval, only the stored keyframes that have cosine similarity above $\tau_{\mathrm{sim}}{=}0.60$ are passed to geometric verification. 


\begin{table}[t]\centering\small\setlength{\tabcolsep}{6pt}\renewcommand{\arraystretch}{1.25}
\caption{Fraction of keyframe queries yielding a valid geometrically verified
loop between AnyLoc (DINOv2-VLAD) and\ DBoW2 BRIEF}
\label{tab:loop_vpr}
\begin{tabular}{l ccc}
\toprule
Dataset & AnyLoc & BRIEF & ratio \\
\midrule
EuRoC          & \best{23.5\%} & 5.7\%  & 4.1$\times$ \\
NTU-VIRAL      & \best{44.7\%} & 22.5\% & 2.0$\times$ \\
Botanic Garden & \best{1.4\%}  & 0.2\%  & 7.0$\times$ \\
SubT-MRS       & \best{9.2\%}  & 4.2\%  & 2.2$\times$ \\
\bottomrule
\end{tabular}
\end{table}

\begin{table}[t]\centering\small\setlength{\tabcolsep}{4.5pt}\renewcommand{\arraystretch}{1.25}
\caption{The AnyLoc cosine similarity retrieval gate impact on positive loop closure.}
\label{tab:loop_gate}
\begin{tabular}{l cc}
\toprule
Dataset & \begin{tabular}{@{}l@{}}Percentage of Keyframe \\ $\ge$ $\tau_{sim}$=0.6\end{tabular} & \begin{tabular}{@{}l@{}}Percentage of Positive \\ Loop Closure\end{tabular} \\
\midrule
EuRoC          & 90\% & 28\%   \\
NTU-VIRAL      & 98\% & 48\%   \\
Botanic Garden & 89\% & 1.6\%  \\
SubT-MRS       & 98\% & 9.6\%  \\
\bottomrule
\end{tabular}
\end{table}

Among $764$ successful runs with valid trajectories including the baseline OKVIS2 and VINS-Fusion, loop closure leaves most sequences nearly unchanged. In $417$ runs, the loop-closed and odometry-only ATE differ by less than 0.05 m or $10\%$. This mainly reflects trajectories with few true revisits, where loop constraints are rarely triggered. Meaningful improvements ($\ge\!0.05$\,m and $\ge\!10\%$) occur in $328$ runs, compared with only $19$ regressions. The gains are concentrated on revisit-rich datasets, affecting $70\%$ of NTU-VIRAL runs ($134/191$), $50\%$ of SubT-MRS runs ($14/28$), and $46\%$ of EuRoC runs ($115/251$). In contrast, only $19$--$25\%$ of Botanic Garden runs improve meaningfully ($36/144$ grayscale, $29/150$ color), as its mostly non-revisiting forest trails provide few loop opportunities. When loop closure is effective, the average ATE reduction is largest on EuRoC ($54\%$), and about $37\%$ on NTU-VIRAL and SubT-MRS.

The two retrieval paths behave differently, as delineated in Table~\ref{tab:loop_vpr}. AnyLoc confirms substantially more loops per keyframe than BRIEF, indicating that the front-end-agnostic DINOv2-VLAD descriptor exposes roughly $2$--$7\times$ more true revisits under its geometric verifier. However, the DINOv2-VLAD cosine score is weak as an absolute gate. As reported in Table~\ref{tab:loop_gate}, most keyframes exceed $\tau_{\mathrm{sim}}{=}0.60$, which rejects only a handful of candidates, yet only $28\%$ of EuRoC, $48\%$ of NTU-VIRAL, $9.6\%$ of SubT-MRS, and $1.6\%$ of Botanic Garden candidates survive geometric verification.

Thus, the cosine similarity gate provides too high a recall but low precision. Candidate pairs in the evaluated dataset lie in the ${\sim}0.5$--$0.7$ similarity range for the same sequence, whereas verified loops cluster much higher, with median scores of $0.81$--$0.90$. The current threshold is therefore too permissive, admitting many low-value candidates that increase verification cost without producing loop constraints. A tighter gate would be preferable, especially on embedded platforms. In practice, the local-descriptor matching and RANSAC stage perform the decisive filtering. Botanic Garden is an extreme case. Repetitive vegetation keeps candidate similarity high (median $0.71$), while confirmed loops reach a median of $0.90$ and only $1.6\%$ are verified, indicating poor global descriptive capability in such a scene.

Loop closure is not uniformly safe. The active regressions cluster in three places. First, OKVIS2's own DBoW2 relocalization accounts for most of them (e.g.\ Botanic gray $1005\_07$ $15.7{\rightarrow}21.4$\,m, $1005\_00$ $5.7{\rightarrow}8.3$\,m), where outdoor vegetation triggers false relocalizations in its bag-of-words map. Second, NTU-VIRAL $\mathrm{nya}\_02$ is a cross-method false-loop hotspot. Most systems regress there, such as SP+LG mono ($0.382{\rightarrow}0.504$\,m) and GFTT+LK mono ($0.425{\rightarrow}0.489$\,m). Third, isolated false closures appear on EuRoC $\mathrm{V}2\_01$ for the LightGlue configurations (SP+LG stereo $0.102{\rightarrow}0.200$\,m, SIFT+LG mono $0.095{\rightarrow}0.181$\,m), where an already-accurate odometry leaves little to gain, and a single incorrect constraint dominates the residual error.

\subsection{Cross-platform analysis}
\label{sec:xplatform}

\begin{figure}[t]
  \includegraphics[width=0.5\textwidth]{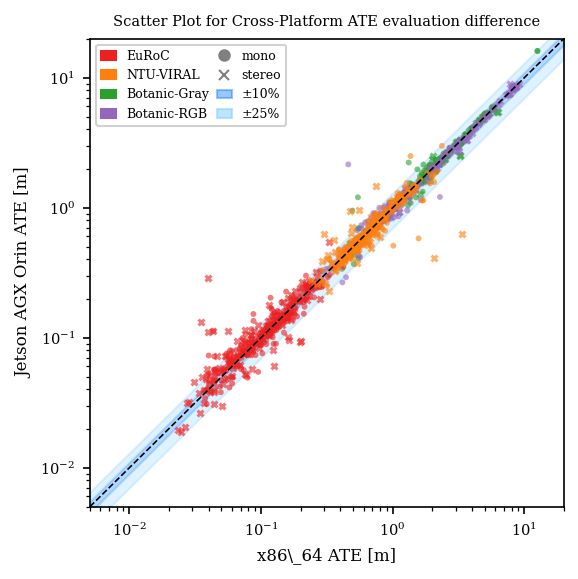}
  \caption{Cross-platform ATE[m] reproducibility scatter plot in log scale. The black dashed line indicates exact parity (Jetson equivalent to x86-64), with shaded bands representing $\pm10\%$ and $\pm25\%$ margins of difference. Data points are categorized by dataset (colors) and sensor configuration (mono: circles, stereo: crosses).}
  \label{fig:xplatform_scatter}
\end{figure}

Both platforms ran with the exact same code, weights, and configuration. ATE should be invariant apart from numerical effects. Fig.~\ref{fig:xplatform_scatter} verifies this by plotting Jetson ATE against x86-64 ATE on log--log axes, where the diagonal denotes identical accuracy. Most points are largely saturated at the diagonal. The median relative difference is only $6\%$, with most runs within the $\pm25\%$ band. 

The remaining scatter is mainly explained by scale. Absolute differences are small, with centimeter-level EuRoC runs showing the largest relative spread, while meter-scale Botanic Garden runs agree within a tighter band. EuRoC is therefore the most sensitive case. A few millimeters of numerical perturbation is significant for a centimeter-level trajectory but negligible for a meter-level one. Moreover, the motion-capture and laser-tracker ground truth is itself only sub-centimeter accurate at best, so a few-millimeter cross-platform difference falls within the ground truth's own uncertainty and is not physically meaningful.

This also explains the apparent mode dependence in Fig.~\ref{fig:xplatform_scatter}. Stereo is not intrinsically less consistent. Overall, it is slightly tighter than monocular ($4.8\%$ vs.\ $7.5\%$ median). The visible stereo scatter is dominated by low-ATE EuRoC stereo runs and a few marginal cases that fail differently across platforms. For example, Ra+LG on NTU-VIRAL $\mathrm{sbs}\_02$ gives $3.4$\,m on x86-64 but $0.6$\,m on Jetson, while Ra+LK on Botanic Gray $1018\_00$ gives $16.0$\,m on x86-64 and $1.2$\,m on Jetson. 

The discrepancy originates primarily in the front-end. Learned extractors are deployed as platform-specific TensorRT FP16 engines, serialized separately for Arm and x86-64. FP16 rounding, target-dependent kernel choices, native floating-point arithmetic, and non-deterministic multi-threading can produce slightly different keypoints and descriptors, which then propagate through tracking and optimization. In the outlier sequences, the per-frame tracking counts are in fact near-identical across platforms (e.g.\ ${\approx}209$ temporal and ${\approx}172$ stereo matches on both for $\mathrm{sbs}\_02$). The difference is that the Jetson triangulates a few more stable landmarks ($+16$ triangulated and $+19$ stereo-anchored on $\mathrm{sbs}\_02$, and ${\approx}{+}4$ each on $1018\_00$). In an already weakly-constrained window, this small surplus is enough to steer the optimizer to the better-conditioned solution, lowering the steady-state cost on $\mathrm{sbs}\_02$ and bounding the velocity estimate on $1018\_00$, whereas x86-64 settles into a higher-residual minimum. Overall, the resulting ATE differences consist of numerical perturbations amplified by the estimator, which are more prevalent, especially on short and highly accurate trajectories.

\subsection{Runtime analysis}
\label{sec:runtime}

\begin{figure*}[t]
  \centering
  \includegraphics[width=\textwidth]{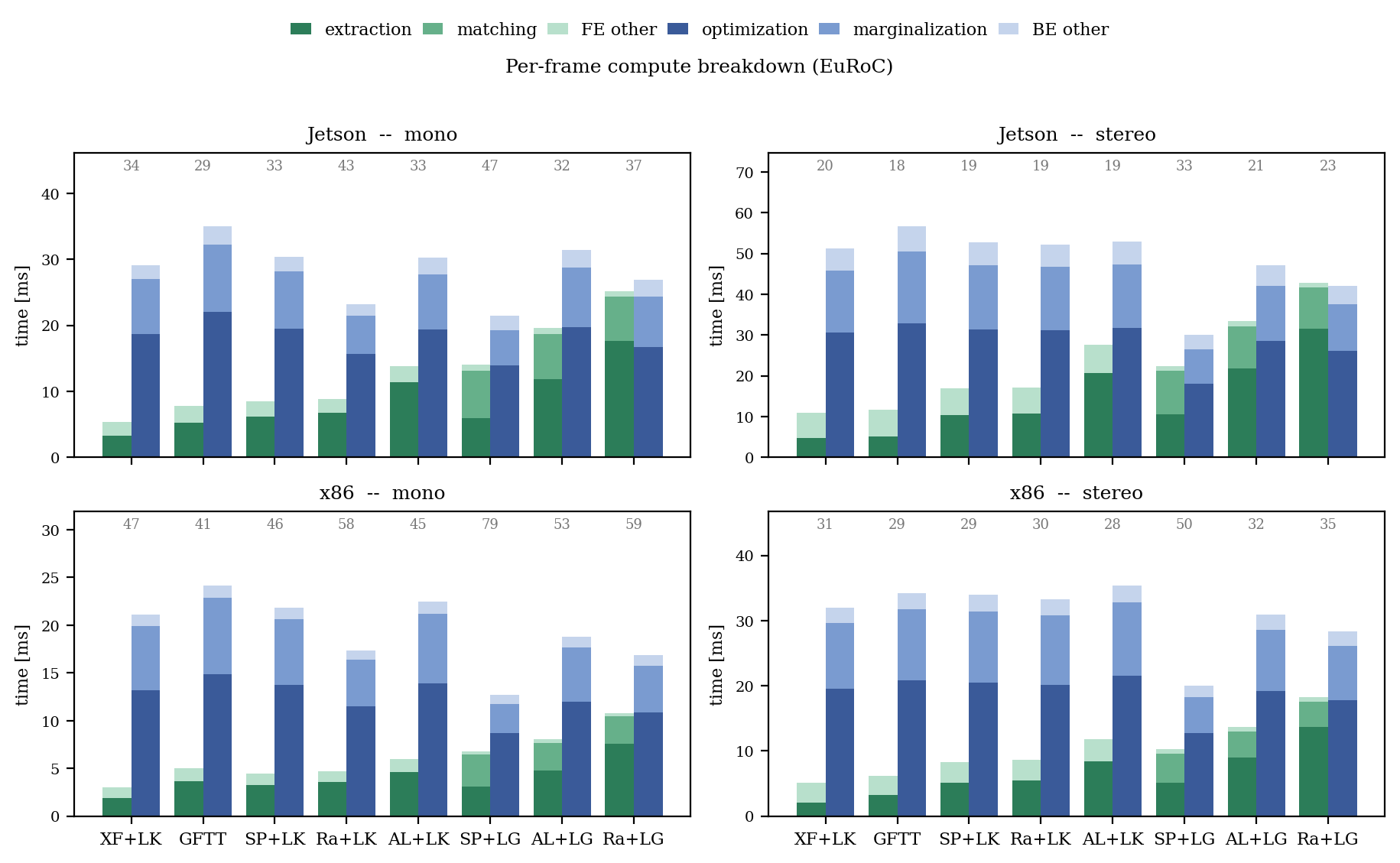}
  \caption{Average per-frame compute breakdown on the EuRoC dataset. The stacked bars illustrate the latency contributions of individual front-end and back-end components across various feature configurations, hardware platforms, and operating modes (mono/stereo). The grey values above each bar indicate the overall system throughput in frames per second (FPS), which is dictated by the slower concurrent stage.}
  \label{fig:runtime}
\end{figure*}

\begin{figure}[t]
  \centering
  \includegraphics[width=0.5\textwidth]{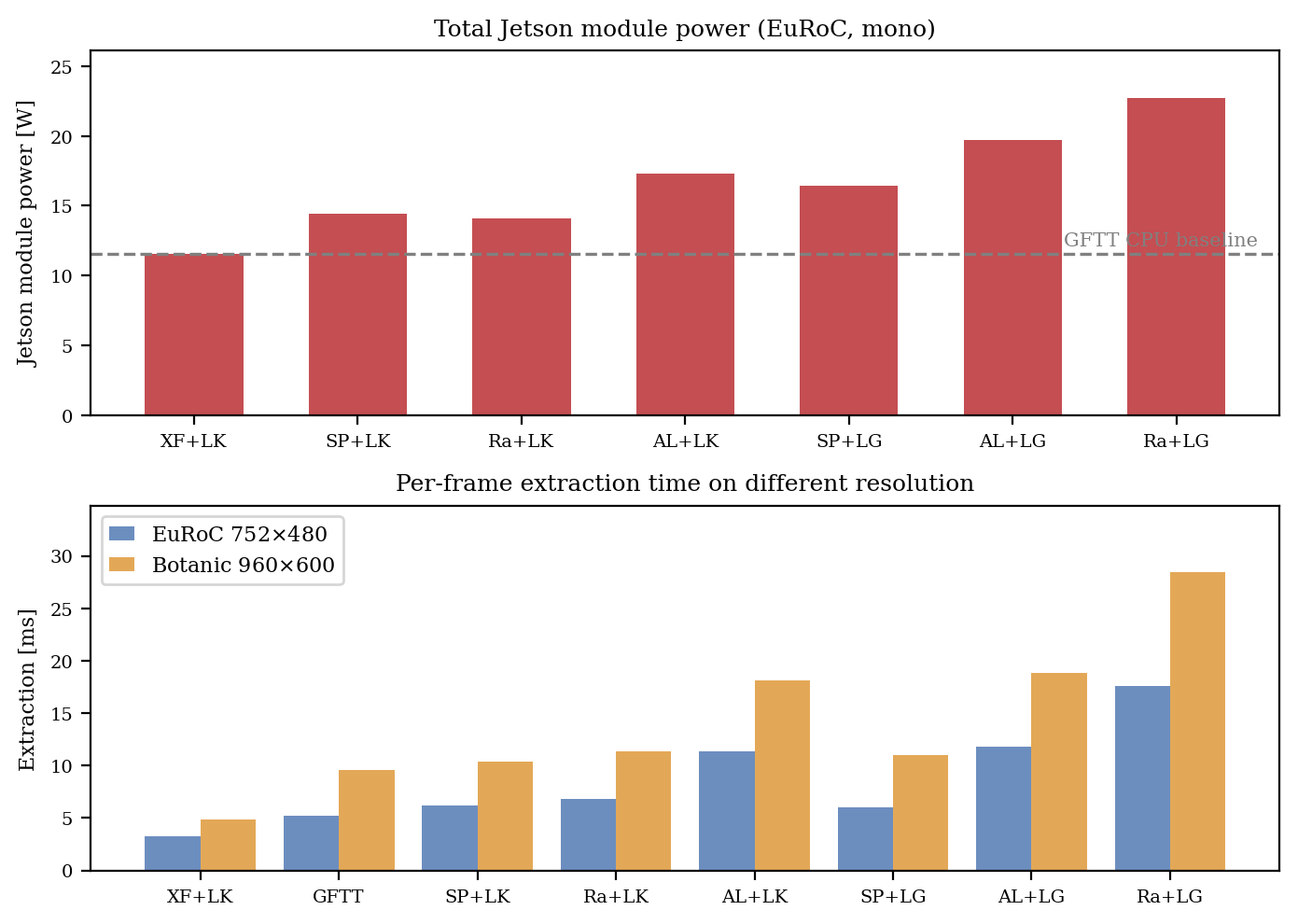}
\caption{(Top) Jetson power consumption and (Bottom) feature-extraction runtime on different resolution input across front-end configurations.}
  \label{fig:runtime_energy}
\end{figure}

We profile the computational and energy footprints on both platforms in Fig. \ref{fig:runtime} and Fig. \ref{fig:runtime_energy}. Because feature tracking (front-end) and sliding-window optimization (back-end) execute concurrently, overall per-frame latency is bound by the slower of the two stages. As in VINS-Fusion, we cap the Ceres solver time at 40 ms. The component breakdown reveals that Ceres optimization dominates back-end compute time, whereas feature extraction is the primary bottleneck within the front-end. Notably, LightGlue introduces a substantial overhead relative to optical flow costs. This explicit matching step is otherwise absent in the optical-flow (LK) pathway.

In monocular mode, the back-end determines overall latency for all configurations, so front-end choice has little effect on system throughput. This is consistent with the VINS-Fusion design, where only every other frame is passed to the estimator, providing additional headroom for the optimizer to converge. Even the heaviest front-end, Ra+LG (25 ms on Jetson mono), remains back-end-bound. Conversely, the lowest monocular throughput occurs with GFTT, not because of front-end cost, but because its saturated keypoint count and high track survival rate increase the number of residuals entering the sliding-window optimization, making the back-end solve more expensive.

Transitioning to stereo mode substantially increases the back-end load due to the introduction of additional stereo residuals. For the front-end, leveraging batch-2 inference for extraction and matching constrains the computational increase for LightGlue configurations to approximately 1.7x, preventing a strict doubling of the processing cost. In this mode, the computational loads of the front-end and back-end stages become more evenly balanced. On the Jetson platform, Ra+LG emerges as the sole front-end-bound configuration in stereo mode, due to the ALIKED descriptor head engine needing to execute sequentially after the RaCo extractor. In contrast, Ra+LK proves to be one of the most lightweight front-ends, while XF+LK registers as the fastest front-end overall.

This back-end bottleneck also applies to the x86-64 workstation. Although the RTX 3080 Ti processes learned front-ends faster than the Jetson GPU, end-to-end throughput only improves by 1.3–1.7x. Because the Ceres back-end remains the limiting factor on both platforms, the front-end's computational headroom is largely unutilized. Consequently, both systems comfortably exceed camera frame rates in the evaluated dataset with TensorRT acceleration (29–47 FPS on Jetson, 41–79 FPS on x86-64). However, the front-ends differ significantly in energy consumption and resolution scaling. Total Jetson module power spans from a baseline of ~11.5 W (GFTT) to 22.7 W for Ra+LG. This ~2x variation is driven entirely by the GPU power required for the LightGlue matcher and the DINOv2 loop-closure engine, while CPU power remains relatively constant.

Furthermore, front-end computational cost scales proportionally with image resolution. Upgrading from EuRoC (752x480) to Botanic (960x600) increases the pixel count by ~1.6x, causing a corresponding linear 1.6x rise in extraction time across all methods. In summary, while learned front-ends effectively hide their latency behind the concurrent back-end optimization, they impose higher GPU energy and memory costs. XF+LG is omitted because its monocular pipeline never initializes, leaving no back-end to time, whereas SIFT is omitted as its CPU runtime is roughly 53 ms, which is almost twice the heaviest learned extractors and would dominate the axes.

\subsection{Keypoint count ablation}
\label{sec:kp_ablation}

To assess the impact of the keypoint limit parameter, we employed two other  values, 192 and 384, on the Jetson AGX Orin, both of which are multiples of 32. This was tested across all nine front-ends in both monocular and stereo configurations on a subset of easy, challenging, earlier identified failure-prone sequences (EuRoC: MH05, V101, V203; NTU-VIRAL: eee\_01, nya\_01, sbs\_02; Botanic Garden Gray: 1005\_01, 1018\_00).

\begin{figure*}[t]
  \centering
  \includegraphics[width=\textwidth]{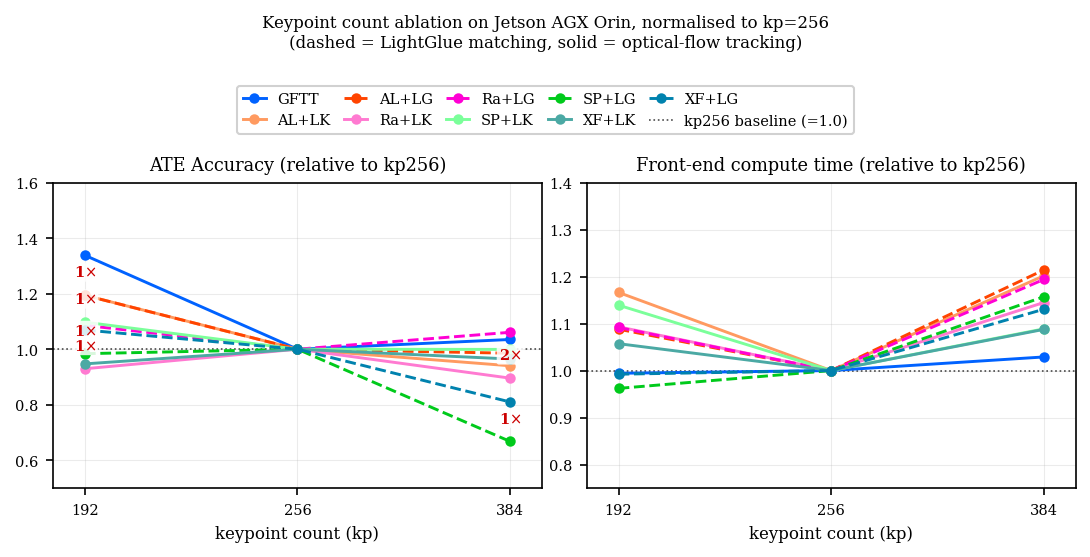}
  \caption{Keypoint count ablation on Jetson AGX Orin. Relative accuracy (left) and front-end compute time (right) are normalized to a 256-keypoint baseline (1.0). Dashed lines indicate LightGlue matching while solid lines denote optical flow. Red crosses with numerical value mark number of tracking failures.}
  \label{fig:kp_ablation}
\end{figure*}

Figure \ref{fig:kp_ablation} summarizes the results, normalizing all data points against the 256-keypoint baseline. The left panel shows the relative ATE, while the right details the mean front-end computation time (extraction and matching). Overall, trajectory accuracy remains largely insensitive to the keypoint budget within this range, though a few configurations reveal specific failure boundaries. Reducing the cap to 192 induces tracking failures for AL+LK and SP+LK (on NTU sbs\_02) and Ra+LK (Botanic 1018\_00 stereo), all of which successfully converge at 256. Increasing the cap to 384 offers negligible accuracy gains while triggering new divergences, such as SP+LG on NTU nya\_01 and Ra+LK on Botanic 1005\_01 mono. The recurring XF+LG failure persists across all sweeps, reflecting an inherent matcher limitation rather than a keypoint capacity constraint.

The front-end runtime shows a non-monotonic dependence on the tracked-keypoint budget. With the exception of SP+LG, the learned front-ends attain a local minimum at 256 keypoints. This optimum is implementation-specific rather than a generic consequence of using a multiple of 32. In our CUDA replacement for ALIKED's DKD stage, fixed 64-thread blocks create a trade-off between exposed parallelism and per-keypoint work. At 192 keypoints, the grid provides less parallel work, so launch overhead, scheduling overhead, and memory-latency effects are less effectively amortized. At 256 keypoints, the kernel has a more favorable occupancy. At 384 keypoints, the kernel remains well occupied, but the larger feature set increases the actual per-keypoint computation and memory traffic, causing runtime to rise. Because RaCo reuses the ALIKED descriptor engine, this runtime minimum applies equally to both ALIKED and RaCo configurations. SP+LG instead relies on its stock decoder and NMS, causing its computational cost to scale linearly with the keypoint budget. The CPU-based GFTT baseline remains expectedly flat.

Ultimately, setting the maximum tracked-keypoint count to 256 is the preferred choice, especially for ALIKED and RaCo, as both sit at the optimal CUDA kernel efficiency point while providing a sufficient feature buffer to prevent the starvation failures observed at lower keypoint counts and avoiding the diminishing accuracy returns and escalating computational costs with higher keypoint counts.

\subsection{RaCo covariance weighting ablation}
\label{sec:raco_cov_ablation}

\begin{table}[t]
  \centering
    \caption{Per-sequence \(\Delta\)ATE (\%) vs.\ the standard Ra+LG baseline (no covariance head) for LightGlue. Negative = lower error; m/s = mono/stereo.}
  \label{tab:cov_ablation}
  \setlength{\tabcolsep}{4pt}
  \begin{tabular}{ll cccc}
    \toprule
    Seq & & $\lambda\!=\!0.25$ & $\lambda\!=\!0.5$ & $\lambda\!=\!0.75$ & $\lambda\!=\!1$ \\
    \midrule
    \multicolumn{6}{l}{\textbf{EuRoC}} \\
    V1\_01 & (m) & -14.6\% & -25.9\% & -24.4\% & -22.6\% \\
    V1\_01 & (s) & -4.6\% & -1.5\% & +15.6\% & -7.3\% \\
    V2\_03 & (m) & +17.0\% & +1.7\% & -2.4\% & -2.0\% \\
    V2\_03 & (s) & +15.9\% & +16.2\% & -11.6\% & -25.4\% \\
    MH\_05 & (m) & +4.0\% & +6.3\% & +5.9\% & +5.8\% \\
    MH\_05 & (s) & -12.1\% & -13.7\% & -16.8\% & -18.1\% \\
    \multicolumn{6}{l}{\textbf{NTU-VIRAL}} \\
    eee\_01 & (m) & -7.5\% & -15.4\% & -13.1\% & -15.0\% \\
    eee\_01 & (s) & -7.9\% & -10.0\% & -6.2\% & -10.8\% \\
    nya\_01 & (m) & +5.6\% & +8.1\% & +10.7\% & +14.4\% \\
    nya\_01 & (s) & \ding{55} & -22.1\% & -16.0\% & -18.3\% \\
    sbs\_02 & (m) & -1.5\% & -5.3\% & -5.6\% & +4.7\% \\
    sbs\_02 & (s) & -80.0\% & +36.0\% & -80.1\% & -80.0\% \\
    \multicolumn{6}{l}{\textbf{Botanic Garden Gray}} \\
    1005\_01 & (m) & +33.7\% & +42.4\% & +47.6\% & +41.3\% \\
    1005\_01 & (s) & -6.2\% & -2.9\% & -0.8\% & +0.5\% \\
    1018\_00 & (m) & -56.6\% & -63.8\% & -65.4\% & -66.5\% \\
    1018\_00 & (s) & +9.7\% & +15.5\% & +17.1\% & +18.5\% \\
    \midrule
    \multicolumn{2}{l}{\textbf{Median}} & -4.6\% & -2.2\% & -5.9\% & -9.1\% \\
    \bottomrule
  \end{tabular}
\end{table}

We sweep the covariance gain $\lambda \in \{0.25, 0.5, 0.75, 1\}$ of Equation~\eqref{eq:our_blend} for Ra+LG on the x86-64 workstation, reusing the keypoint-ablation sequences (Sec.~\ref{sec:kp_ablation}) in both monocular and stereo-inertial modes. We report the ATE change relative to the standard Ra+LG baseline, where the covariance head is disabled. Table~\ref{tab:cov_ablation} lists the per-sequence deltas. One NTU-VIRAL stereo run ($\lambda=0.25$) diverged and is excluded from the median.

The sweep is meaningful only for LightGlue, because the two front-ends deliver covariance to the back-end very differently. LightGlue re-extracts and re-matches features every frame, so every live observation carries a fresh per-keypoint $\Sigma_n$ and the reweighting is pervasive. Optical flow instead produces a covariance only when a keypoint is first detected and thereafter tracks it geometrically. Once a track is established, its subsequent observations revert to the isotropic weight of Equation~\eqref{eq:our_sqrtinfo}. Since steady-state tracking is dominated by long-lived tracks, almost no $\Sigma_n$ ever reaches the back-end and the sweep is effectively useless for Ra+LK. We therefore report Ra+LG only.

Covariance weighting helps almost exclusively in stereo. The aggregate median at $\lambda=1$ is $-9.1\%$, but it splits sharply by mode. It is marginally worse by $+1.4\%$ in monocular against $-14.4\%$ in stereo. The learned weight is attached to the left-camera observation, and the back-end applies it both to that keypoint's temporal reprojection factor and to its additional cross-camera stereo factor. Where metric depth is observable, this anisotropic Mahalanobis weighting (Equation~\eqref{eq:our_sqrtinfo}) down-weights poorly localized detections and better conditions the stereo bundle adjustment. In monocular mode, the same rescaling merely perturbs an already scale-ambiguous problem (Appendix~\ref{app:cov}), so its net effect is neutral to slightly harmful. Across the sweep, full reliance on the RaCo covariance ($\lambda=1$) is the most helpful setting.

The per-sequence deltas provide a more detailed picture than the aggregate medians. The largest single gain is a rescue. On NTU sbs\_02 stereo, where plain Ra+LG is a documented failure ($3.357$~m, Sec.~\ref{sec:accuracy analysis}), covariance weighting cuts ATE by ${\sim}80\%$. Botanic Garden Gray $1018\_00$ sequence reverses the story as covariance weighting improves the monocular trajectory by up to $-66.5\%$ ($\lambda=1$), yet degrades the stereo trajectory by $+18.5\%$ because the limiting factor there is absent disparity in the open vegetation scene that reweighting cannot overcome. On the other Botanic Garden Gray $1005\_01$ sequence, covariance instead clearly hurts in monocular ($+41.3\%$). Intermediate weighting is not the safest as NTU nya\_01 stereo diverges at $\lambda=0.25$ yet recovers for every $\lambda\geq0.5$, as a partially anisotropic $\Sigma_{\mathrm{eff}}$ can ill-condition the sliding-window Hessian at specific operating points.

This pattern is expected. MAC-VO~\cite{macvo} gains reliably because its uncertainty head is trained end-to-end with the negative-log-likelihood loss of Equation~\eqref{eq:macvo_nll}, calibrating $\hat{\Sigma}_f$ against matching error inside the estimator. RaCo's head is trained out of context, so only its relative anisotropy is trustworthy, and a global gain $\lambda$ is needed to rescale it. The static gain we tested is, at best, a proxy. We foresee that a learned, online estimator of $\lambda$, analogous to MAC-VO's jointly trained head, would be the natural next step for RaCo to improve the estimator.

\section{Conclusion}
\label{sec:conclusion}

This work presented DL-VINS-Factory, a modular VI-SLAM framework for evaluating learned visual front-ends under a common tightly coupled back-end, loop-closure pipeline, and hardware deployment stack. By combining ALIKED, RaCo, SuperPoint, and XFeat with either optical-flow tracking or LightGlue matching, the framework isolates the effect of front-end design while preserving the estimator and geometric verification stages.

The evaluation shows that no single front-end dominates across all environments. Learned matching is beneficial under large viewpoint changes, as observed in aerial and aggressive-motion sequences, but optical-flow tracking remains highly competitive in structured scenes and more reliable in low-parallax or visually degraded settings. Stereo improves accuracy when sufficient, well-conditioned disparity is available, but can degrade performance in vegetation-heavy scenes where stereo correspondences are sparse or weakly constrained. These results indicate that front-end representation alone does not determine VI-SLAM performance. Scene geometry, motion excitation, feature survival, and back-end conditioning are equally important.

Loop closure further improves accuracy, mainly on revisit-rich trajectories. AnyLoc provides a front-end-agnostic retrieval path and exposes more valid loop candidates than the BRIEF-based baseline, but its cosine similarity is too permissive as a standalone gate. In practice, local descriptor matching and RANSAC remain essential for rejecting false positives, especially in natural environments.

Finally, the TensorRT hardware acceleration demonstrates that learned front-ends can run in real time on embedded hardware. Other ablation studies that are still unaddressed include the TensorRT quantization effect (INT8, BF16, FP32) on accuracy and throughput, as well as the different ALIKED and SuperPoint variants. Since the sliding-window optimizer is often the throughput bottleneck, front-end latency is partly hidden by concurrent execution. However, learned extraction, LightGlue matching, and DINOv2-based retrieval increase GPU power and memory demand. Overall, learned visual front-ends are practical for embedded VI-SLAM, but their advantage is conditional rather than universal and requires careful selection for the intended deployment area.


\bibliographystyle{IEEEtran}
\bibliography{IEEEabrv,References}

\appendices

\clearpage
\newpage

\section{Covariance-Weighted Reprojection}
\label{app:cov}

This section details the covariance formulation proposed in Sec.~\ref{sec:learned_front}. Given measurement uncertainty from a front-end network head (RaCo), we fold it into the optimizer through Jacobian-based covariance propagation, following the metrics-aware weighting of MAC-VO~\cite{macvo}. We first summarize the MAC-VO model that inspired our weighting, then describe our adaptation to the sliding-window reprojection back-end.

\subsection{MAC-VO covariance model}
MAC-VO learns a per-match $2$D matching uncertainty and propagates it to a $3$D landmark covariance that weights pose-graph optimization. Alongside the optical flow $\hat{\mathbf{f}}\in\mathbb{R}^2$, its matching network predicts a diagonal pixel-space covariance $\hat{\Sigma}_f=\mathrm{diag}(\sigma_u^2,\sigma_v^2)$, trained with the negative-log-likelihood loss
\begin{equation}
  \mathcal{L}_{\mathrm{cov}} = \sum_i \alpha_i\Bigl( (\mathbf{y}-\hat{\mathbf{f}}_i)^{\!\top} \hat{\Sigma}_f^{-1}(\mathbf{y}-\hat{\mathbf{f}}_i) + \log\det\hat{\Sigma}_f \Bigr),
  \label{eq:macvo_nll}
\end{equation}
where $\mathbf{y}$ is the ground-truth flow and $\alpha_i$ weights the $i$-th decoder iteration.

Given the pinhole model $(f_x,f_y,c_x,c_y)$, a pixel $(u,v)$ at depth $z$ back-projects to 3D coordinate $\mathbf{x}=\bigl((u-c_x)z/f_x,\,(v-c_y)z/f_y,\,z\bigr)$. Propagating the pixel uncertainty $(\sigma_u^2,\sigma_v^2)$ and the depth uncertainty $ \sigma_z^2$ through this map yields the diagonal terms
\begin{equation}
  \begin{aligned}
    \sigma_x^2 &= \bigl(\sigma_u^2 \sigma_z^2 + \sigma_u^2 z^2 + (u-c_x)^2 \sigma_z^2\bigr)/f_x^2,\\
    \sigma_y^2 &= \bigl(\sigma_v^2 \sigma_z^2 + \sigma_v^2 z^2 + (v-c_y)^2 \sigma_z^2\bigr)/f_y^2,\\
  \end{aligned}
  \label{eq:macvo_diag}
\end{equation}
and, because the three coordinates share the common depth factor, the off-diagonal inter-axis terms
\begin{equation}
  \begin{aligned}
    \sigma_{xz}&= \sigma_z^2(u-c_x)/f_x,\\
    \sigma_{yz}&= \sigma_z^2(v-c_y)/f_y,\\
    \sigma_{xy}&= \sigma_z^2(u-c_x)(v-c_y)/(f_x f_y),
  \end{aligned}
  \label{eq:macvo_offdiag}
\end{equation}
assembled into the full $3\times3$ landmark covariance ${}^{c}\Sigma$. The relative pose is then estimated by minimizing the Mahalanobis distance between registered $3$D keypoints,
\begin{equation}
  \begin{aligned}
    T^\star&=\arg\min_{T_t}\sum_i\bigl\|\mathbf{x}_{i,t-1}-T_t\,{}^{c}\mathbf{x}_{i,t}\bigr\|_{\Sigma_i}^2,\\
    \Sigma_i&={}^{c}\Sigma_{i,t-1} + \mathbf{R}_t\,{}^{c}\Sigma_{i,t}\,\mathbf{R}_t^{\!\top},
  \end{aligned}
  \label{eq:macvo_pgo}
\end{equation}
solved by Levenberg--Marquardt, so low-confidence matches contribute less to the estimated motion.

\subsection{Adaptation in DL-VINS-Factory}
Our setting differs in two respects. First, we do not train an end-to-end flow-uncertainty head. Instead RaCo's covariance estimator predicts, in the pre-processed (letterboxed) image space, a per-keypoint $2\times2$ pixel covariance $\Sigma_{\mathrm{px}}=LL^{\!\top}$ from a lower-triangular Cholesky factor $L$; we rescale $\Sigma_{\mathrm{px}}$ to native pixel coordinates by the squared letterbox factor, the same scale that maps the keypoints back to native coordinates. Second, our back-end minimizes a $2$D reprojection error in the sliding window rather than registering $3$D points. We therefore stop the propagation at the \emph{reprojection plane} instead of forming the full $3\times3$ covariance of~\eqref{eq:macvo_diag}--\eqref{eq:macvo_offdiag}, which keeps the modification confined to the per-observation weight. The pixel covariance is mapped onto the normalized image plane, on which the reprojection residual is defined, by the Jacobian of the camera un-projection $\pi^{-1}$ evaluated at the keypoint $\mathbf{p}$,
\begin{equation}
  \Sigma_n = J_{\pi^{-1}}\,\Sigma_{\mathrm{px}}\,J_{\pi^{-1}}^{\!\top},\qquad
  J_{\pi^{-1}} = \left.\frac{\partial \pi^{-1}}{\partial \mathbf{p}}\right|_{\mathbf{p}}.
  \label{eq:our_transform}
\end{equation}
For a pinhole model this is $\mathrm{diag}(1/f_x,1/f_y)$; our implementation uses a single shared focal length ($f_x\!=\!f_y$), and the same construction extends to the omnidirectional and equidistant models supported by our camera library through their analytic unprojection Jacobians. Let $\mathbf{r}\in\mathbb{R}^2$ be the reprojection residual on this plane. VINS-Fusion weights every such residual with a fixed isotropic square-root information $\Omega_0=\sigma_0^{-1}I$, a single reprojection-noise scalar $\sigma_0$ shared by all keypoints. We keep this Ceres square-root-information convention but replace the shared scalar with a per-keypoint anisotropic factor,
\begin{equation}
  \mathbf{r}'=\Omega\,\mathbf{r},\qquad \Omega=\mathrm{chol}\bigl(\Sigma_{\mathrm{eff}}^{-1}\bigr),
  \label{eq:our_sqrtinfo}
\end{equation}
so that $\|\mathbf{r}'\|_2^2=\mathbf{r}^{\!\top}\Sigma_{\mathrm{eff}}^{-1}\mathbf{r}$ is a Mahalanobis distance and each visual factor contributes a Hessian block $J_{\mathbf{r}}^{\!\top}\Sigma_{\mathrm{eff}}^{-1}J_{\mathbf{r}}$, with $J_{\mathbf{r}}$ the Jacobian of $\mathbf{r}$ with respect to the optimization states. This is the same inverse-covariance--weighted normal-equation \emph{structure} as MAC-VO's $3$D term in~\eqref{eq:macvo_pgo}, here applied at the $2$D reprojection plane rather than to registered $3$D points. Setting $\Sigma_{\mathrm{eff}}=\sigma_0^2 I$ recovers the VINS-Fusion weight exactly.

RaCo predicts covariance in its own training context rather than in the context of our estimator, so the absolute scale of $\Sigma_n$ is not directly reliable even though its relative anisotropy is informative. We therefore do not use $\Sigma_n$ at face value but rescale its information by a single global gain $\lambda$:
\begin{equation}
  \Sigma_{\mathrm{eff}}^{-1}=\lambda\,\Sigma_n^{-1},\qquad \lambda\in(0,1],
  \label{eq:our_blend}
\end{equation}
so that $\lambda=1$ applies RaCo's covariance at its native scale while $\lambda<1$ uniformly inflates it, softening the learned weight toward a weaker but still anisotropic contribution. Well-conditioning is maintained by a lower clamp on the gain and, optionally, a small additive floor on $\Sigma_{\mathrm{px}}$, with the isotropic $\sigma_0^{-1}I$ retained as a fallback whenever the Cholesky factorization fails. The isotropic VINS-Fusion weight is therefore \emph{not} recovered as $\lambda\!\to\!0$; it is instead supplied by a separate baseline in which the covariance head is disabled, and that disabled-head baseline, rather than any $\lambda$ setting, is the reference against which the ablation is measured. We sweep $\lambda$ without tuning on the evaluation data and report the sensitivity of trajectory accuracy to it.

\section{Per-Sequence Accuracy Results}
\label{app:ate}

This section presents the per-sequence ATE results and 3D trajectory plot for each evaluated dataset in Sec. V-A.

\begin{table*}[t]\centering\scriptsize\setlength{\tabcolsep}{1.0pt}\renewcommand{\arraystretch}{1.2}
\begin{threeparttable}
\caption{EuRoC ATE RMSE [m] measured on x86\_64 platform.}
\label{tab:euroc_ate}
\begin{tabular}{l *{12}{c}}\toprule
Method & MH01 & MH02 & MH03 & MH04 & MH05 & V101 & V102 & V103 & V201 & V202 & V203 & Avg \\ \midrule
\multicolumn{13}{@{}l}{\textbf{Mono-Inertial}}\\[1pt]
VINS-Fusion & \sbest{0.186}/0.159 & 0.088/0.096${^*}$ & 0.135/0.103 & \sbest{0.206}/0.211${^*}$ & 0.400/0.341 & \best{0.057}/0.059${^*}$ & 0.086/0.077 & 0.160/0.138 & 0.059/0.082${^*}$ & \ding{55} & 0.205/0.143 & 0.158$^{\dagger}$/0.141$^{\dagger}$ \\
OKVIS2 & 0.245/0.087 & \best{0.074}/0.058 & 0.197/\sbest{0.058} & 0.307/0.204 & 0.321/\best{0.092} & \sbest{0.065}/\best{0.032} & \best{0.076}/\best{0.028} & 0.110/\best{0.039} & 0.133/0.074 & 0.110/0.064 & 0.146/\best{0.067} & 0.162/\sbest{0.073} \\
\cmidrule(lr){1-13}
GFTT (LK) & 0.232/0.069 & 0.137/0.052 & 0.214/0.071 & \best{0.176}/\sbest{0.123} & 0.293/0.223 & 0.108/0.067 & 0.125/0.050 & 0.127/0.078 & 0.060/0.044 & \sbest{0.088}/0.070 & 0.131/\sbest{0.076} & \sbest{0.154}/0.084 \\
SIFT (LK) & 0.265/0.072 & 0.110/0.049 & 0.203/\best{0.057} & 0.217/0.129 & 0.415/0.177 & 0.095/\sbest{0.036} & 0.093/0.043 & 0.137/0.139${^*}$ & 0.091/0.064 & 0.126/0.117 & \sbest{0.118}/0.100 & 0.170/0.089 \\
SIFT (LG) & 0.251/0.072 & 0.430/0.097 & 2.331/0.098 & 0.425/0.242 & 1.060/0.161 & \ding{55} & 0.323/0.085 & 0.147/0.083 & 0.095/0.181${^*}$ & 0.160/0.159 & 0.412/0.326 & 0.563$^{\dagger}$/0.141 \\
AL (LK) & 0.188/0.159 & 0.153/0.147 & 0.168/0.119 & 0.273/0.254 & 0.386/0.359 & 0.135/0.133 & 0.118/0.100 & 0.130/0.122 & 0.079/0.065 & 0.126/0.120 & 0.175/0.172 & 0.176/0.159 \\
AL (LG) & 0.218/\best{0.061} & 0.098/0.053 & \best{0.121}/0.060 & 0.253/\best{0.121} & \best{0.225}/\sbest{0.110} & 0.097/0.038 & 0.099/0.045 & \sbest{0.085}/\sbest{0.055} & \best{0.058}/0.060${^*}$ & 0.094/\best{0.040} & 0.253/0.132 & \best{0.146}/\best{0.070} \\
Ra (LK) & 0.219/0.075 & 0.158/0.057 & 0.159/0.066 & 0.219/0.169 & 0.422/0.131 & 0.099/0.037 & 0.116/0.047 & 0.167/0.087 & 0.058/0.057 & 0.089/0.049 & 0.122/0.105 & 0.166/0.080 \\
Ra (LG) & \best{0.179}/0.095 & \sbest{0.086}/\best{0.040} & \sbest{0.134}/0.071 & 0.360/0.258 & 0.388/0.122 & 0.071/0.039 & \sbest{0.079}/\sbest{0.042} & 0.091/0.060 & 0.059/0.045 & 0.119/0.074 & 0.278/0.109 & 0.168/0.087 \\
SP (LK) & 0.225/\sbest{0.065} & 0.133/\sbest{0.044} & 0.183/0.073 & 0.232/0.167 & 0.332/0.155 & 0.104/0.045 & 0.103/0.049 & 0.164/0.072 & \sbest{0.058}/\best{0.039} & 0.130/0.069 & \best{0.118}/0.077 & 0.162/0.078 \\
SP (LG) & 0.206/0.115 & 0.180/0.066 & 0.230/0.063 & 0.281/0.164 & \sbest{0.265}/0.175 & 0.099/0.040 & 0.112/0.049 & \best{0.073}/0.102${^*}$ & 0.061/0.079${^*}$ & 0.134/0.070 & 0.289/0.149 & 0.176/0.098 \\
XF (LK) & 0.211/0.091 & 0.161/0.060 & 0.208/0.109 & 0.219/0.178 & 0.322/0.243 & 0.100/0.045 & 0.085/0.044 & 0.143/0.080 & 0.068/\sbest{0.044} & \best{0.087}/\sbest{0.043} & 0.127/0.085 & 0.157/0.093 \\
XF (LG) & \ding{55} & \ding{55} & \ding{55} & \ding{55} & \ding{55} & \ding{55} & \ding{55} & \ding{55} & \ding{55} & \ding{55} & \ding{55} & \ding{55} \\
\midrule
\multicolumn{13}{@{}l}{\textbf{Stereo-Inertial}}\\[1pt]
VINS-Fusion & 0.253/0.222 & 0.217/0.198 & 0.326/0.211 & 0.434/0.392 & 0.314/0.329${^*}$ & 0.108/0.074 & 0.108/0.099 & 0.112/0.113${^*}$ & 0.132/0.081 & 0.124/0.112 & 0.318/0.205 & 0.222/0.185 \\
OKVIS2 & \best{0.053}/\best{0.022} & \sbest{0.042}/\best{0.022} & \sbest{0.110}/\best{0.037} & \best{0.099}/\best{0.066} & \best{0.142}/\best{0.076} & \best{0.019}/\best{0.014} & \best{0.050}/\best{0.016} & \best{0.041}/\best{0.022} & \best{0.027}/\best{0.019} & \best{0.037}/\best{0.016} & \best{0.044}/\best{0.019} & \best{0.060}/\best{0.030} \\
\cmidrule(lr){1-13}
GFTT (LK) & \sbest{0.082}/0.041 & 0.080/0.037 & 0.123/0.061 & 0.179/0.157 & 0.182/0.140 & \sbest{0.040}/0.025 & 0.129/0.061 & 0.129/0.061 & 0.077/0.044 & 0.146/0.055 & 0.174/\sbest{0.063} & \sbest{0.122}/0.068 \\
AL (LK) & 0.085/0.044 & 0.139/0.126 & 0.160/0.074 & 0.181/0.222${^*}$ & 0.261/0.258 & 0.056/0.051 & 0.088/0.066 & 0.113/0.100 & 0.073/0.062 & 0.155/0.158${^*}$ & 0.233/0.197 & 0.140/0.123 \\
AL (LG) & 0.096/0.044 & 0.088/0.034 & 0.142/\sbest{0.039} & \sbest{0.144}/0.146${^*}$ & \sbest{0.148}/0.116 & 0.057/\sbest{0.023} & 0.111/0.035 & 0.114/0.048 & 0.100/0.052 & \sbest{0.112}/0.056 & 0.281/0.105 & 0.127/\sbest{0.063} \\
Ra (LK) & 0.083/0.075 & 0.161/0.047 & 0.162/0.095 & 0.283/0.207 & 0.219/0.238${^*}$ & 0.045/0.076${^*}$ & 0.112/0.096 & 0.118/0.076 & 0.084/0.043 & 0.156/0.144 & \sbest{0.163}/0.123 & 0.144/0.111 \\
Ra (LG) & 0.085/0.076 & \best{0.040}/0.035 & 0.114/0.053 & 0.222/0.222${^*}$ & 0.265/\sbest{0.112} & 0.041/0.027 & \sbest{0.084}/\sbest{0.028} & 0.098/\sbest{0.039} & 0.075/0.050 & 0.125/\sbest{0.054} & 0.845/0.209 & 0.181/0.082 \\
SP (LK) & 0.117/0.056 & 0.081/0.114${^*}$ & 0.139/0.087 & 0.188/0.169 & 0.225/0.197 & 0.053/0.050 & 0.109/0.092 & 0.114/0.103 & \sbest{0.069}/0.038 & 0.140/0.130 & 0.188/0.106 & 0.129/0.104 \\
SP (LG) & 0.083/\sbest{0.031} & 0.069/\sbest{0.034} & \best{0.109}/0.043 & 0.169/\sbest{0.125} & 0.155/0.120 & 0.040/0.027 & 0.098/0.039 & \sbest{0.073}/0.060 & 0.102/0.200${^*}$ & 0.153/0.065 & 0.325/0.115 & 0.125/0.078 \\
XF (LK) & 0.129/0.053 & 0.120/0.124${^*}$ & 0.159/0.076 & 0.201/0.194 & 0.232/0.204 & 0.073/0.058 & 0.129/0.108 & 0.135/0.086 & 0.077/\sbest{0.037} & 0.156/0.138 & 0.224/0.163 & 0.149/0.113 \\
XF (LG) & 0.613/0.606 & 0.366/0.369${^*}$ & 0.288/0.290${^*}$ & 0.412/0.183 & 0.428/0.298 & 0.102/0.098 & 0.124/0.092 & 0.296/0.294 & 0.282/0.273 & 0.513/0.511 & 0.725/0.724 & 0.377/0.340 \\
\bottomrule
\end{tabular}
\atenotesref
\end{threeparttable}\end{table*}

\begin{table*}[t]\centering\scriptsize\setlength{\tabcolsep}{1.0pt}\renewcommand{\arraystretch}{1.2}
\begin{threeparttable}
\caption{NTU-VIRAL ATE RMSE [m] measured on x86\_64 platform.}
\label{tab:ntu_viral_ate}
\begin{tabular}{l *{10}{c}}\toprule
Method & EEE01 & EEE02 & EEE03 & NYA01 & NYA02 & NYA03 & SBS01 & SBS02 & SBS03 & Avg \\ \midrule
\multicolumn{11}{@{}l}{\textbf{Mono-Inertial}}\\[1pt]
VINS-Fusion & \ding{55} & \ding{55} & \ding{55} & \best{0.577}/\best{0.363} & 0.491/0.604${^*}$ & 1.352/\sbest{0.679} & \ding{55} & \ding{55} & 1.312/0.771 & 0.933$^{\dagger}$/0.604$^{\dagger}$ \\
OKVIS2 & \ding{55} & \ding{55} & \ding{55} & \ding{55} & \ding{55} & \ding{55} & \ding{55} & \ding{55} & \ding{55} & \ding{55} \\
\cmidrule(lr){1-11}
GFTT (LK) & 1.396/0.826 & 1.065/0.583 & 1.304/0.875 & 1.571/1.010 & 0.425/0.489${^*}$ & \sbest{1.203}/0.703 & 1.824/0.906 & 1.120/0.535 & 1.430/0.451 & 1.260/0.709 \\
SIFT (LK) & 1.613/0.830 & 1.019/0.457 & 1.389/0.976 & 2.189/0.609 & 0.460/0.526${^*}$ & 1.482/0.918 & 1.750/1.421 & 1.743/1.314 & 1.641/0.591 & 1.476/0.849 \\
SIFT (LG) & 1.536/1.167 & 0.610/\best{0.275} & 1.179/1.117 & 2.726/0.920 & 0.865/0.600 & \ding{55} & 1.452/1.209 & 1.317/1.074 & \best{0.677}/0.487 & 1.295$^{\dagger}$/0.856$^{\dagger}$ \\
AL (LK) & 1.674/1.708${^*}$ & 1.145/0.990 & 1.141/1.033 & 0.679/0.486 & 0.449/0.452${^*}$ & 1.307/0.946 & 2.169/2.030 & 1.974/2.029${^*}$ & 2.064/1.917 & 1.400/1.288 \\
AL (LG) & \sbest{0.848}/\best{0.466} & 0.776/\sbest{0.293} & \sbest{0.969}/0.758 & 1.200/0.464 & \sbest{0.397}/\best{0.345} & 1.314/\best{0.650} & 1.465/0.989 & \best{0.997}/\sbest{0.530} & 0.840/0.403 & 0.978/\sbest{0.544} \\
Ra (LK) & 1.178/0.759 & 0.987/0.445 & 1.263/\sbest{0.729} & 0.765/0.517 & 0.410/0.457${^*}$ & 1.342/0.839 & 1.461/\sbest{0.759} & 1.416/0.593 & 1.173/0.491 & 1.110/0.621 \\
Ra (LG) & \best{0.793}/\sbest{0.521} & \best{0.423}/0.318 & \best{0.843}/\best{0.389} & 0.916/0.576 & 0.403/0.478${^*}$ & 1.318/0.680 & \sbest{1.241}/\best{0.714} & 1.121/0.609 & \sbest{0.836}/\best{0.331} & \best{0.877}/\best{0.513} \\
SP (LK) & 1.276/0.735 & 1.111/0.561 & 1.319/0.838 & \sbest{0.643}/0.504 & 0.411/\sbest{0.408} & 1.307/0.720 & 2.362/1.366 & \sbest{1.099}/\best{0.447} & 1.020/0.540 & 1.172/0.680 \\
SP (LG) & 1.616/0.861 & \sbest{0.600}/0.415 & 1.123/0.744 & 0.766/0.495 & \best{0.382}/0.504${^*}$ & \best{1.108}/0.760 & \best{0.960}/0.792 & 1.167/0.665 & 1.032/\sbest{0.378} & \sbest{0.973}/0.624 \\
XF (LK) & 0.923/0.789 & 1.013/0.616 & 1.243/1.093 & 0.904/\sbest{0.426} & 0.463/0.415 & 1.330/0.718 & \ding{55} & 1.117/0.897 & 1.515/0.807 & 1.063$^{\dagger}$/0.720$^{\dagger}$ \\
XF (LG) & \ding{55} & \ding{55} & \ding{55} & \ding{55} & \ding{55} & \ding{55} & \ding{55} & \ding{55} & \ding{55} & \ding{55} \\
\midrule
\multicolumn{11}{@{}l}{\textbf{Stereo-Inertial}}\\[1pt]
VINS-Fusion & 0.589/0.581 & \sbest{0.606}/\best{0.254} & 0.605/0.405 & 0.605/\sbest{0.384} & 0.550/0.559${^*}$ & 0.844/0.484 & 0.570/0.580${^*}$ & \sbest{0.573}/0.480 & 0.895/0.646 & 0.649/0.486 \\
OKVIS2 & 2.449/0.596 & 0.761/0.301 & 1.140/1.257${^*}$ & \best{0.439}/\best{0.353} & 0.640/0.855${^*}$ & 2.290/2.272 & 0.976/0.497 & 3.559/2.308 & 0.816/0.868${^*}$ & 1.452/1.034 \\
\cmidrule(lr){1-11}
GFTT (LK) & 0.683/0.359 & 0.666/\sbest{0.266} & 0.791/0.331 & 0.590/0.523 & 0.547/0.458 & 0.771/\best{0.442} & \sbest{0.455}/0.394 & 0.713/0.417 & 1.182/0.423 & 0.711/\sbest{0.402} \\
AL (LK) & 0.870/0.854 & 0.665/0.610 & 0.645/0.548 & \sbest{0.476}/0.575${^*}$ & \best{0.403}/\sbest{0.424}${^*}$ & \best{0.687}/0.493 & 0.565/0.550 & 0.644/0.622 & 1.164/1.142 & 0.680/0.647 \\
AL (LG) & \best{0.561}/0.394 & 0.674/0.306 & \best{0.388}/\best{0.284} & 0.507/0.430 & 0.496/\best{0.377} & 0.801/0.499 & 0.464/\sbest{0.365} & \best{0.472}/\best{0.287} & 0.600/\best{0.239} & \best{0.552}/\best{0.354} \\
Ra (LK) & 0.744/0.741 & \best{0.543}/0.524 & 0.557/0.489 & 0.802/0.680 & 0.526/0.458 & 0.752/0.559 & 0.861/0.841 & 0.724/0.698 & 1.470/1.462 & 0.775/0.717 \\
Ra (LG) & \sbest{0.570}/\sbest{0.339} & 1.000/0.602 & 0.525/\sbest{0.289} & 0.528/0.522 & 0.478/0.483${^*}$ & 0.720/\sbest{0.478} & \best{0.447}/0.376 & 3.357/2.074 & \best{0.488}/0.411 & 0.902/0.619 \\
SP (LK) & 0.783/0.763 & 0.760/0.587 & 0.582/0.518 & 0.569/0.543 & 0.466/0.493${^*}$ & \sbest{0.719}/0.544 & 0.677/0.650 & 0.742/0.725 & 1.071/1.003 & 0.708/0.647 \\
SP (LG) & 0.714/\best{0.300} & 0.779/0.364 & \sbest{0.413}/0.318 & 0.648/0.553 & 0.523/0.465 & 0.758/0.559 & 0.482/\best{0.320} & 0.641/\sbest{0.392} & \sbest{0.515}/\sbest{0.343} & \sbest{0.608}/0.402 \\
XF (LK) & 0.724/0.701 & 0.633/0.539 & 0.717/0.645 & 0.686/0.652 & \sbest{0.461}/0.463${^*}$ & 0.968/0.683 & 0.742/0.710 & 0.767/0.754 & 0.910/0.850 & 0.734/0.666 \\
XF (LG) & 4.788/4.786 & 3.688/3.683 & 1.832/1.849${^*}$ & 1.754/1.756${^*}$ & 1.476/1.466 & 1.817/1.815 & 2.295/2.292 & 1.173/1.166 & 2.009/2.018${^*}$ & 2.315/2.315 \\
\bottomrule
\end{tabular}
\atenotesref
\end{threeparttable}\end{table*}

\begin{table*}[t]\centering\scriptsize\setlength{\tabcolsep}{1.0pt}\renewcommand{\arraystretch}{1.2}
\begin{threeparttable}
\caption{Botanic-Garden Gray ATE RMSE [m] measured on x86\_64 platform.}
\label{tab:botanic_gray_ate}
\begin{tabular}{l *{8}{c}}\toprule
Method & 1005-00 & 1005-01 & 1005-07 & 1006-01 & 1008-03 & 1018-00 & 1018-13 & Avg \\ \midrule
\multicolumn{9}{@{}l}{\textbf{Mono-Inertial}}\\[1pt]
VINS-Fusion & \best{1.338}/\best{1.338} & 1.134/1.134 & 2.117/2.117 & 1.189/1.020 & \ding{55} & 0.622/0.622 & 1.368/1.368 & 1.295$^{\dagger}$/1.267$^{\dagger}$ \\
OKVIS2 & 5.650/8.314${^*}$ & 2.523/1.700 & 15.744/21.378${^*}$ & 6.901/2.019 & 8.151/6.288 & 0.710/1.159${^*}$ & 5.561/4.713 & 6.463/6.510${^*}$ \\
\cmidrule(lr){1-9}
GFTT (LK) & 1.800/1.667 & \sbest{0.915}/\sbest{0.842} & 1.891/1.668 & 1.226/0.749 & 2.260/1.958 & 0.741/0.703 & 0.803/0.773 & 1.377/1.194 \\
SIFT (LK) & 2.310/2.078 & 1.230/1.105 & 2.920/2.664 & 1.263/0.614 & 2.386/2.343 & 0.735/0.689 & 1.083/1.083 & 1.704/1.511 \\
SIFT (LG) & 4.197/4.112 & 5.875/5.857 & \ding{55} & \ding{55} & 4.111/4.105 & \ding{55} & \ding{55} & 4.728$^{\dagger}$/4.691$^{\dagger}$ \\
AL (LK) & 1.709/1.537 & 1.369/1.306 & 1.690/1.501 & \sbest{0.952}/0.911 & 2.045/1.972 & 0.554/0.544 & 1.092/1.031 & 1.344/1.257 \\
AL (LG) & 2.313/2.272 & 1.899/1.893 & 1.956/1.931 & 1.850/0.688 & \best{1.743}/1.355 & 0.576/0.543 & 4.020/4.000 & 2.051/1.812 \\
Ra (LK) & 2.131/1.909 & \best{0.751}/\best{0.703} & 1.148/1.054 & \best{0.909}/\sbest{0.430} & \sbest{1.765}/\best{1.158} & \sbest{0.544}/\sbest{0.494} & \best{0.514}/\best{0.470} & \sbest{1.109}/\sbest{0.888} \\
Ra (LG) & 1.815/1.784 & 1.355/1.366${^*}$ & 1.385/1.350 & 1.683/0.617 & 1.790/1.250 & 0.834/0.631 & 1.824/1.800 & 1.526/1.257 \\
SP (LK) & \sbest{1.511}/\sbest{1.352} & 1.050/0.979 & \best{1.087}/\best{0.959} & 0.968/\best{0.413} & 1.852/\sbest{1.174} & \best{0.529}/\best{0.489} & 0.597/0.586 & \best{1.085}/\best{0.850} \\
SP (LG) & 2.159/2.133 & 2.126/2.110 & 12.588/12.550 & 2.321/1.322 & 2.522/1.985 & \ding{55} & \ding{55} & 4.343$^{\dagger}$/4.020$^{\dagger}$ \\
XF (LK) & 1.744/1.610 & \ding{55} & \sbest{1.126}/\sbest{1.017} & 0.993/0.469 & 1.796/1.717 & 0.549/0.506 & \sbest{0.538}/\sbest{0.477} & 1.124$^{\dagger}$/0.966$^{\dagger}$ \\
XF (LG) & \ding{55} & \ding{55} & \ding{55} & \ding{55} & \ding{55} & \ding{55} & \ding{55} & \ding{55} \\
\midrule
\multicolumn{9}{@{}l}{\textbf{Stereo-Inertial}}\\[1pt]
VINS-Fusion & 3.958/3.958 & 2.706/2.706 & 4.465/4.465 & 2.806/2.882${^*}$ & 2.484/2.484 & 0.634/0.634 & 1.252/1.252 & 2.615/2.626${^*}$ \\
OKVIS2 & 3.146/3.132 & 2.777/2.826${^*}$ & \sbest{3.255}/\sbest{3.208} & 2.944/3.286${^*}$ & 2.390/2.503${^*}$ & 0.784/0.775 & 1.069/1.095${^*}$ & 2.338/2.404${^*}$ \\
\cmidrule(lr){1-9}
GFTT (LK) & 3.959/3.818 & 2.462/2.237 & 4.751/4.506 & 3.308/1.903 & 3.138/2.348 & 0.584/\sbest{0.556} & \sbest{1.028}/\sbest{0.932} & 2.747/2.329 \\
AL (LK) & 5.002/5.037${^*}$ & 2.336/2.323 & 4.793/4.616 & 2.735/2.742${^*}$ & 3.017/2.930 & 0.902/0.897 & 1.632/1.710${^*}$ & 2.917/2.893 \\
AL (LG) & \sbest{2.700}/\sbest{2.618} & \best{1.875}/\best{1.819} & \best{3.062}/\best{3.036} & \best{1.875}/\best{1.633} & 2.355/1.917 & \sbest{0.570}/0.557 & 1.316/1.292 & \best{1.965}/\best{1.839} \\
Ra (LK) & 5.017/5.079${^*}$ & 2.529/2.472 & 4.490/4.275 & 3.216/3.236${^*}$ & 3.082/3.026 & \ding{55} & 1.270/1.311${^*}$ & 3.267$^{\dagger}$/3.233$^{\dagger}$ \\
Ra (LG) & 3.183/3.099 & \sbest{2.111}/\sbest{2.067} & 3.672/3.626 & \sbest{2.001}/1.934 & \best{2.140}/\sbest{1.838} & 0.596/0.575 & \best{0.911}/\best{0.886} & \sbest{2.088}/\sbest{2.004} \\
SP (LK) & 4.907/4.847 & 3.095/3.046 & 5.706/5.430 & 3.246/3.252${^*}$ & 3.232/3.181 & 1.273/1.275${^*}$ & 1.590/1.636${^*}$ & 3.293/3.238 \\
SP (LG) & \best{2.039}/\best{1.983} & 3.272/3.240 & \ding{55} & 2.508/\sbest{1.740} & \sbest{2.253}/\best{1.618} & \best{0.533}/\best{0.506} & 6.306/6.245 & 2.818$^{\dagger}$/2.555$^{\dagger}$ \\
XF (LK) & 5.218/5.231${^*}$ & 3.600/3.552 & 5.772/5.669 & 2.532/2.533${^*}$ & 3.463/3.460 & 1.415/1.417${^*}$ & 1.967/2.065${^*}$ & 3.424/3.418 \\
XF (LG) & \ding{55} & \ding{55} & \ding{55} & \ding{55} & \ding{55} & \ding{55} & \ding{55} & \ding{55} \\
\bottomrule
\end{tabular}
\atenotesref
\end{threeparttable}\end{table*}

\begin{table*}[t]\centering\scriptsize\setlength{\tabcolsep}{1.0pt}\renewcommand{\arraystretch}{1.2}
\begin{threeparttable}
\caption{Botanic-Garden RGB ATE RMSE [m] measured on x86\_64 platform.}
\label{tab:botanic_rgb_ate}
\begin{tabular}{l *{8}{c}}\toprule
Method & 1005-00 & 1005-01 & 1005-07 & 1006-01 & 1008-03 & 1018-00 & 1018-13 & Avg \\ \midrule
\multicolumn{9}{@{}l}{\textbf{Mono-Inertial}}\\[1pt]
VINS-Fusion & 0.906/0.906 & 1.275/1.275 & 1.884/1.884 & 1.369/1.407${^*}$ & 1.456/1.456 & 0.786/0.786 & 1.419/1.419 & 1.299/1.305${^*}$ \\
OKVIS2 & 2.141/2.154${^*}$ & 3.713/3.770${^*}$ & 4.875/4.558 & 4.067/5.800${^*}$ & 3.043/4.683${^*}$ & 1.746/1.025 & 5.698/4.359 & 3.612/3.764${^*}$ \\
\cmidrule(lr){1-9}
GFTT (LK) & 1.110/1.102 & 1.004/0.974 & 1.444/1.374 & 1.511/0.943 & 1.595/1.570 & 0.673/0.636 & 1.065/1.065${^*}$ & 1.200/1.095 \\
SIFT (LK) & 1.299/1.255 & 1.131/1.037 & 2.208/1.991 & 1.913/1.897 & 1.942/1.913 & 0.769/0.726 & 1.989/1.961 & 1.607/1.540 \\
SIFT (LG) & \ding{55} & 4.090/4.076 & \ding{55} & \ding{55} & 1.534/1.520 & \sbest{0.359}/\sbest{0.356} & \ding{55} & 1.994$^{\dagger}$/1.984$^{\dagger}$ \\
AL (LK) & 2.282/1.634 & 1.023/0.959 & \sbest{1.224}/1.143 & 1.131/1.100 & 1.320/1.296 & 0.464/0.456 & 0.953/0.941 & 1.200/1.076 \\
AL (LG) & \sbest{0.515}/\sbest{0.508} & 1.138/1.118 & 1.992/1.966 & 1.983/0.580 & \best{1.007}/\best{1.002} & \best{0.314}/\best{0.293} & 2.708/2.700 & 1.380/1.167 \\
Ra (LK) & 0.553/0.565${^*}$ & \best{0.728}/\best{0.685} & \best{1.032}/\best{0.970} & \best{0.868}/\sbest{0.441} & 1.156/1.140 & 0.578/0.538 & \best{0.452}/\best{0.430} & \best{0.767}/\best{0.681} \\
Ra (LG) & \best{0.459}/\best{0.482}${^*}$ & 1.352/1.351 & 1.596/1.527 & 1.256/0.615 & \sbest{1.034}/\sbest{1.003} & 0.627/0.536 & 1.342/1.312 & 1.095/0.975 \\
SP (LK) & 0.806/0.795 & 1.331/1.197 & 1.274/1.191 & \sbest{0.930}/\best{0.335} & 1.350/1.315 & 0.442/0.415 & \sbest{0.648}/\sbest{0.619} & 0.969/\sbest{0.838} \\
SP (LG) & 0.807/0.804 & 1.363/1.350 & 3.262/3.217 & 2.423/1.135 & 1.075/1.072 & 0.532/0.511 & \ding{55} & 1.577$^{\dagger}$/1.348$^{\dagger}$ \\
XF (LK) & 0.868/0.858 & \sbest{0.792}/\sbest{0.712} & 1.288/\sbest{1.103} & 1.089/1.059 & 1.292/1.273 & 0.543/0.516 & 0.721/0.658 & \sbest{0.942}/0.883 \\
XF (LG) & \ding{55} & \ding{55} & \ding{55} & \ding{55} & \ding{55} & \ding{55} & \ding{55} & \ding{55} \\
\midrule
\multicolumn{9}{@{}l}{\textbf{Stereo-Inertial}}\\[1pt]
VINS-Fusion & 1.138/1.138 & \sbest{1.814}/\sbest{1.814} & 4.492/4.492 & 2.684/2.314 & 7.725/7.725 & 0.685/0.685 & 0.847/0.847 & \sbest{2.769}/2.716 \\
OKVIS2 & \sbest{0.902}/0.882 & \best{1.493}/\best{1.564}${^*}$ & \best{1.789}/\best{1.890}${^*}$ & \best{1.876}/\best{1.162} & \best{4.090}/\best{4.598}${^*}$ & 0.690/0.647 & 0.880/0.763 & \best{1.674}/\best{1.644} \\
\cmidrule(lr){1-9}
GFTT (LK) & 1.808/1.303 & 3.035/2.984 & 5.208/4.953 & 3.135/3.049 & 8.656/8.592 & 0.805/0.797 & 0.905/0.914${^*}$ & 3.365/3.227 \\
AL (LK) & 1.232/0.941 & 2.255/1.931 & 4.758/4.463 & 3.516/3.532${^*}$ & 8.166/8.014 & 0.866/0.872${^*}$ & 1.107/1.165${^*}$ & 3.129/2.988 \\
AL (LG) & \best{0.688}/\best{0.643} & 2.416/2.247 & 5.264/5.150 & 4.214/\sbest{1.285} & 8.494/8.233 & \sbest{0.640}/\sbest{0.639} & \sbest{0.572}/\sbest{0.569} & 3.184/\sbest{2.681} \\
Ra (LK) & 1.471/1.157 & 2.575/2.244 & 5.782/5.421 & \sbest{2.597}/2.598${^*}$ & 8.563/8.416 & 0.954/0.972${^*}$ & 1.015/1.050${^*}$ & 3.280/3.123 \\
Ra (LG) & 1.011/\sbest{0.877} & 2.078/1.946 & 6.544/6.290 & 4.046/1.476 & 8.544/9.139${^*}$ & 0.686/0.695${^*}$ & 0.739/0.744${^*}$ & 3.378/3.024 \\
SP (LK) & 1.485/1.132 & 2.158/1.869 & 4.443/4.222 & 3.599/3.649${^*}$ & 8.125/7.988 & 0.776/0.789${^*}$ & 0.953/1.003${^*}$ & 3.077/2.950 \\
SP (LG) & 1.064/0.962 & 3.135/2.859 & 7.838/7.693 & 5.938/1.319 & 8.000/7.745 & \best{0.567}/\best{0.560} & \best{0.544}/\best{0.539} & 3.869/3.097 \\
XF (LK) & 1.397/0.972 & 2.410/1.982 & \sbest{3.994}/\sbest{3.794} & 3.178/3.148 & \sbest{7.461}/\sbest{7.277} & 1.121/1.119 & 1.100/1.152${^*}$ & 2.951/2.778 \\
XF (LG) & 1.924/1.897 & \ding{55} & \ding{55} & \ding{55} & \ding{55} & \ding{55} & \ding{55} & 1.924$^{\dagger}$/1.897$^{\dagger}$ \\
\bottomrule
\end{tabular}
\atenotesref
\end{threeparttable}\end{table*}

\clearpage
\newpage

\begin{figure*}[t]
  \includegraphics[width=1.0\textwidth]{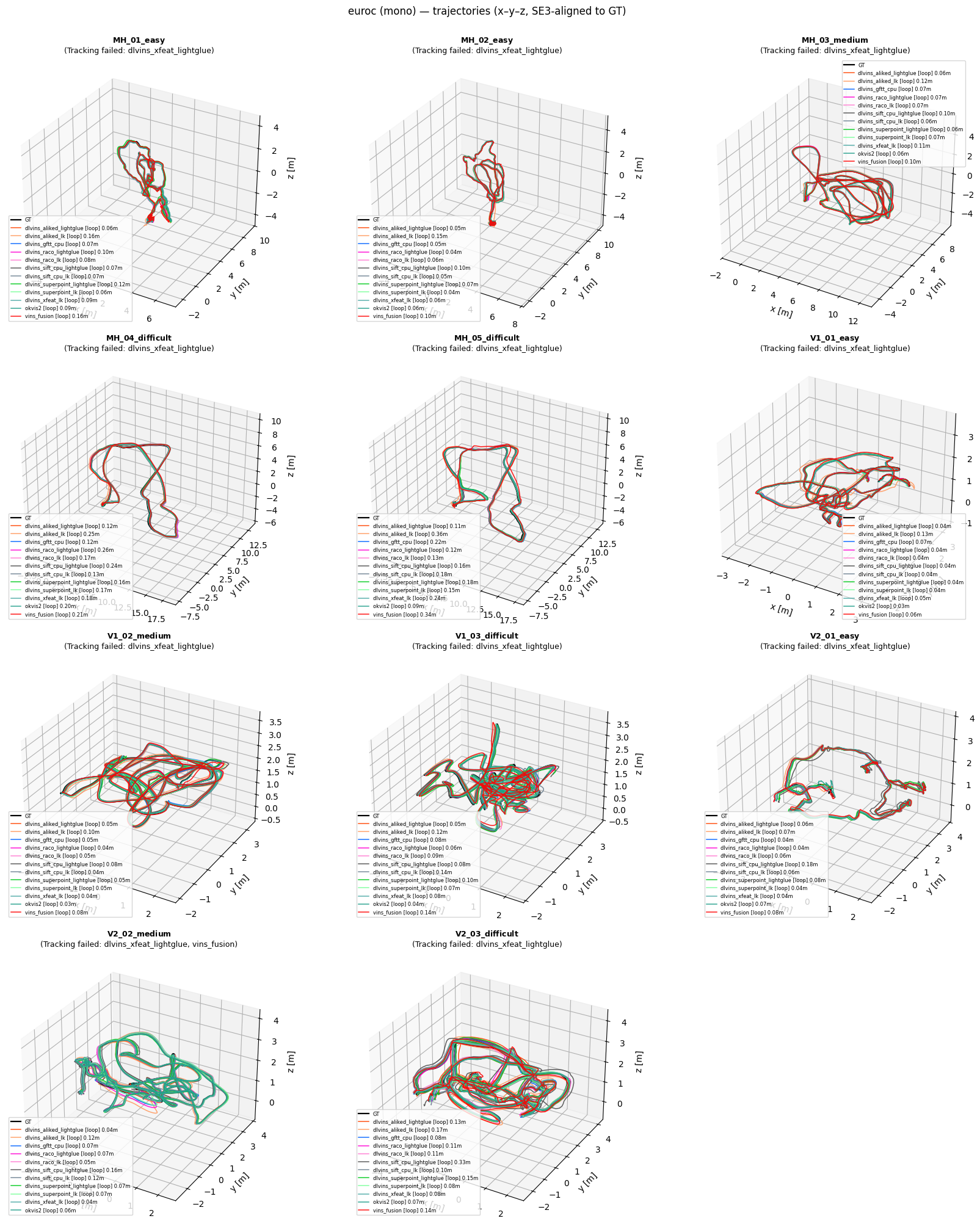}
  \caption{\textbf{EuRoC} mono-inertial 3D trajectory plot.}
  \label{fig:euroc_mono}
\end{figure*}

\begin{figure*}[t]
  \includegraphics[width=1.0\textwidth]{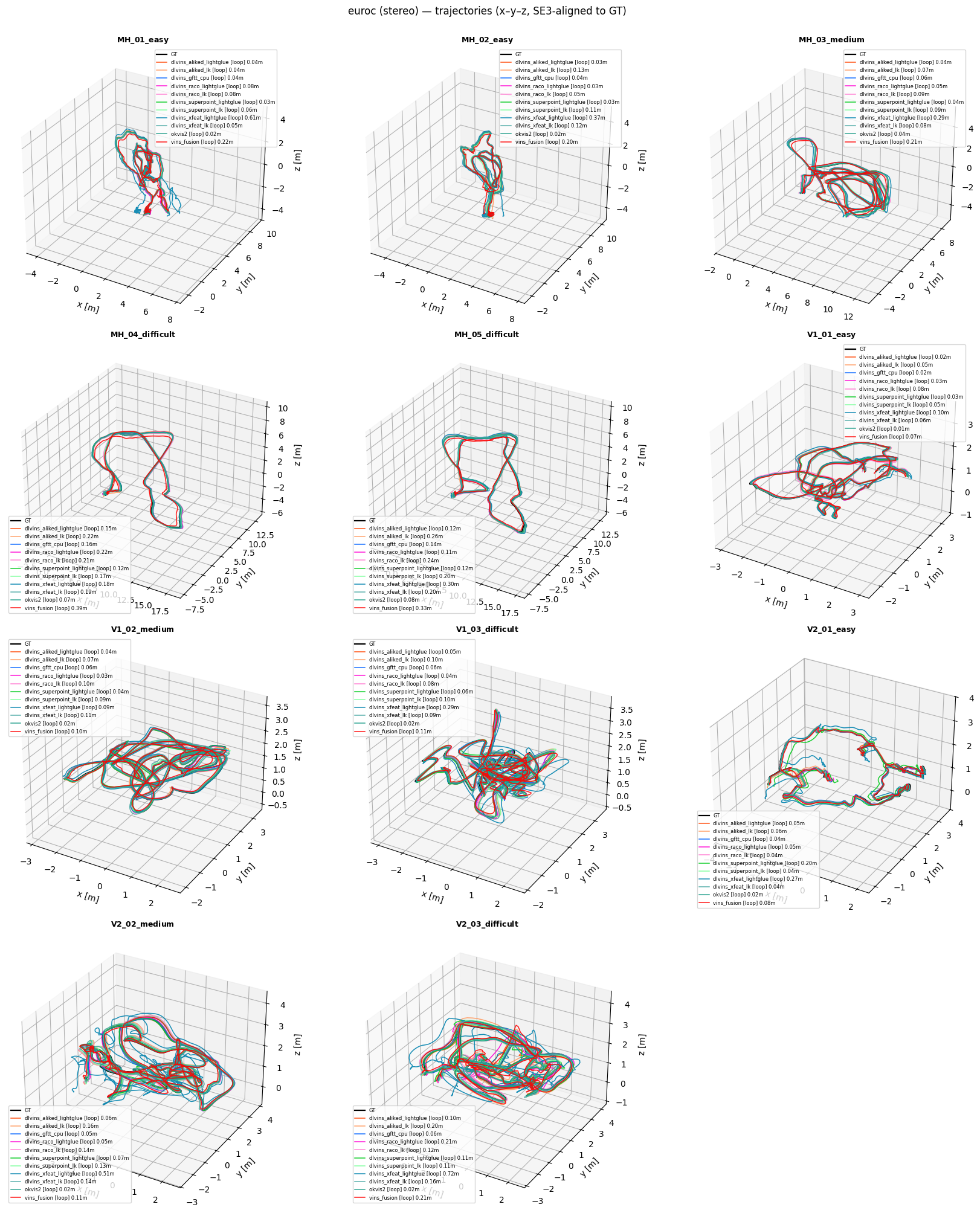}
  \caption{\textbf{EuRoC} stereo-inertial 3D trajectory plot.}
  \label{fig:euroc_stereo}
\end{figure*}

\begin{figure*}[t]
  \includegraphics[width=1.0\textwidth]{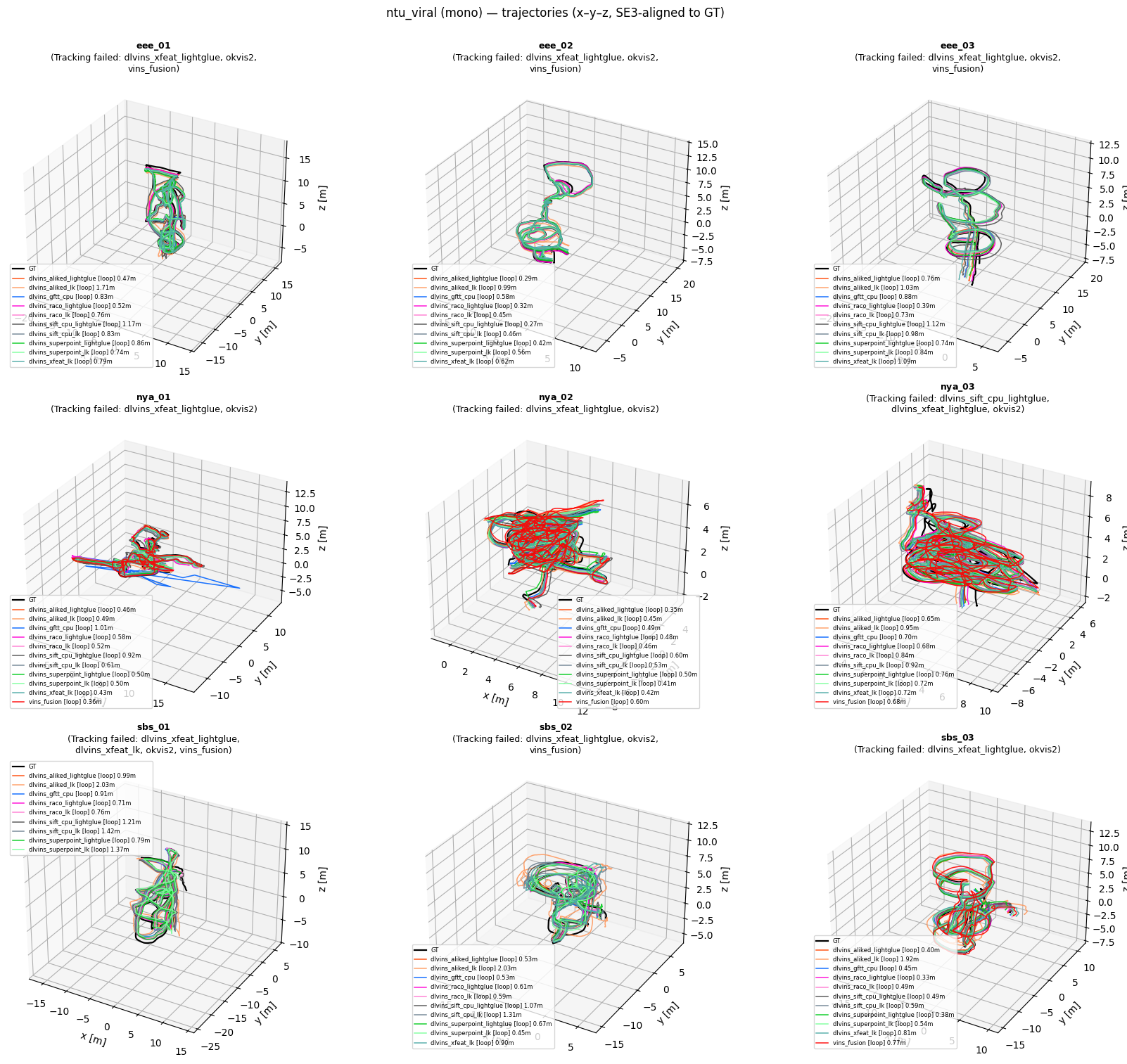}
  \caption{\textbf{NTU-VIRAL} mono-inertial 3D trajectory plot.}
  \label{fig:ntu_mono}
\end{figure*}

\begin{figure*}[t]
  \includegraphics[width=1.0\textwidth]{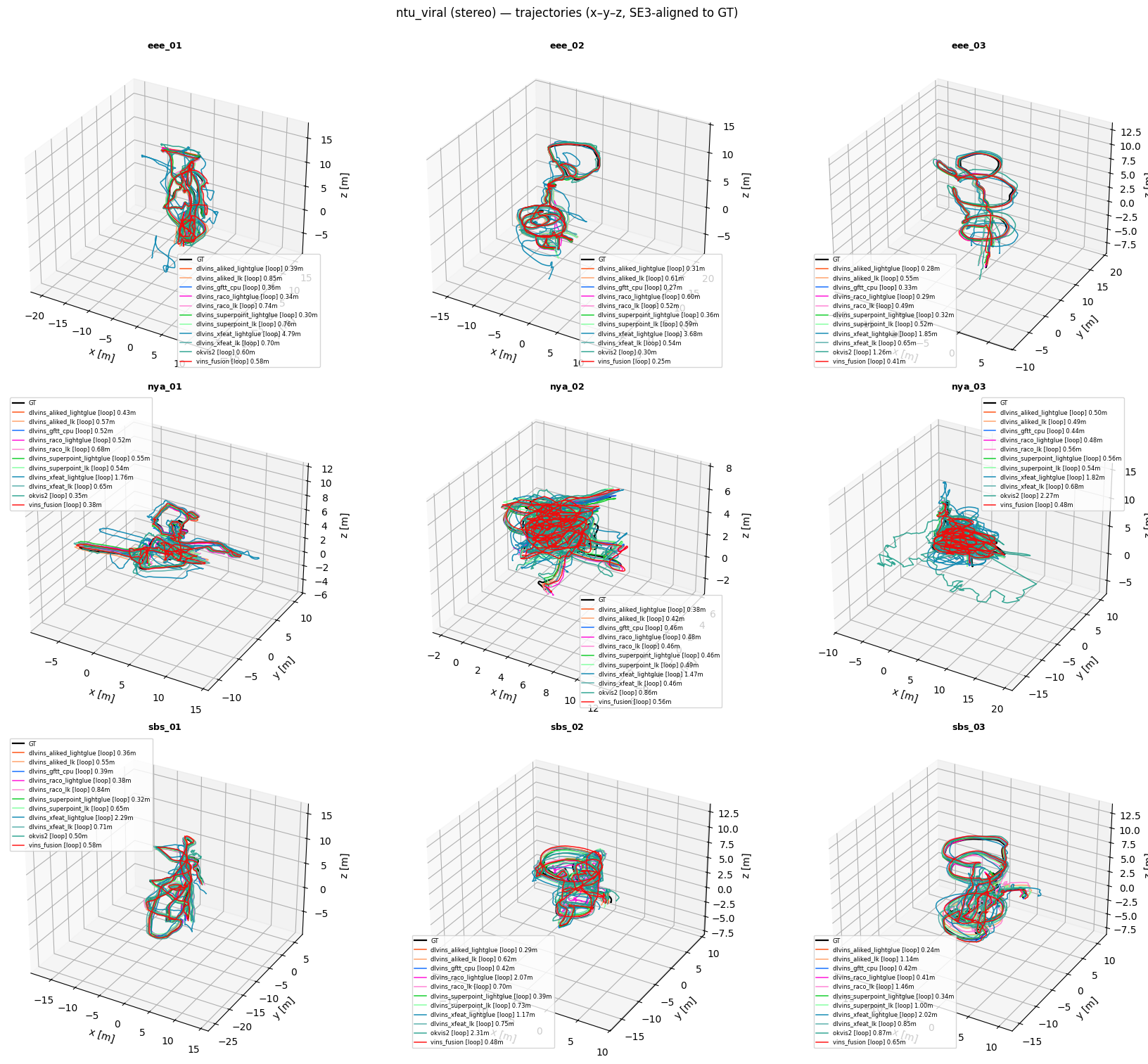}
  \caption{\textbf{NTU-VIRAL} stereo-inertial 3D trajectory plot.}
  \label{fig:ntu_stereo}
\end{figure*}

\begin{figure*}[t]
  \includegraphics[width=1.0\textwidth]{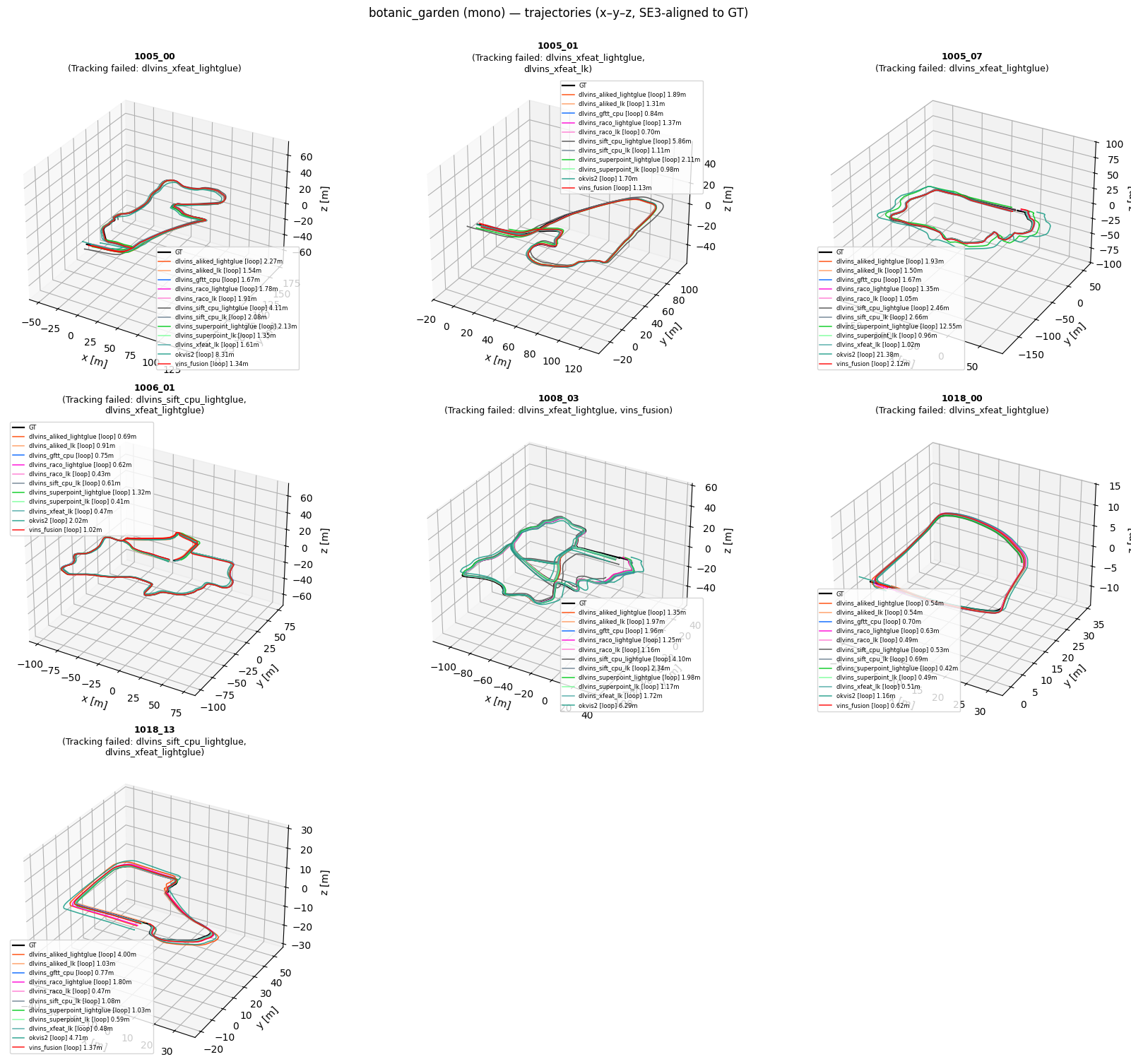}
  \caption{\textbf{Botanic Garden Gray} mono-inertial 3D trajectory plot.}
  \label{fig:bg_g_mono}
\end{figure*}

\begin{figure*}[t]
  \includegraphics[width=1.0\textwidth]{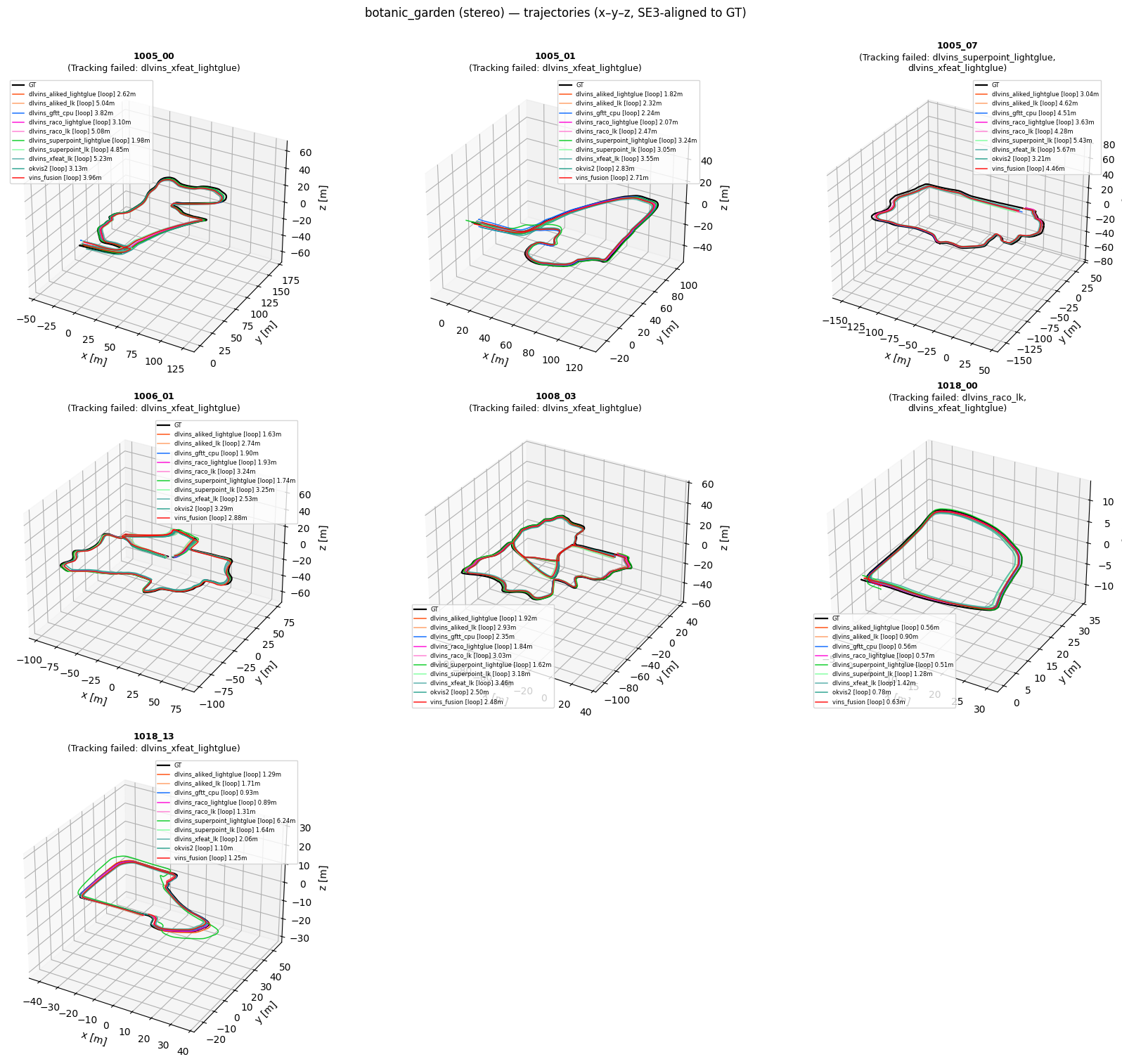}
  \caption{\textbf{Botanic Garden Gray} stereo-inertial 3D trajectory plot.}
  \label{fig:bg_g_stereo}
\end{figure*}

\begin{figure*}[t]
  \includegraphics[width=1.0\textwidth]{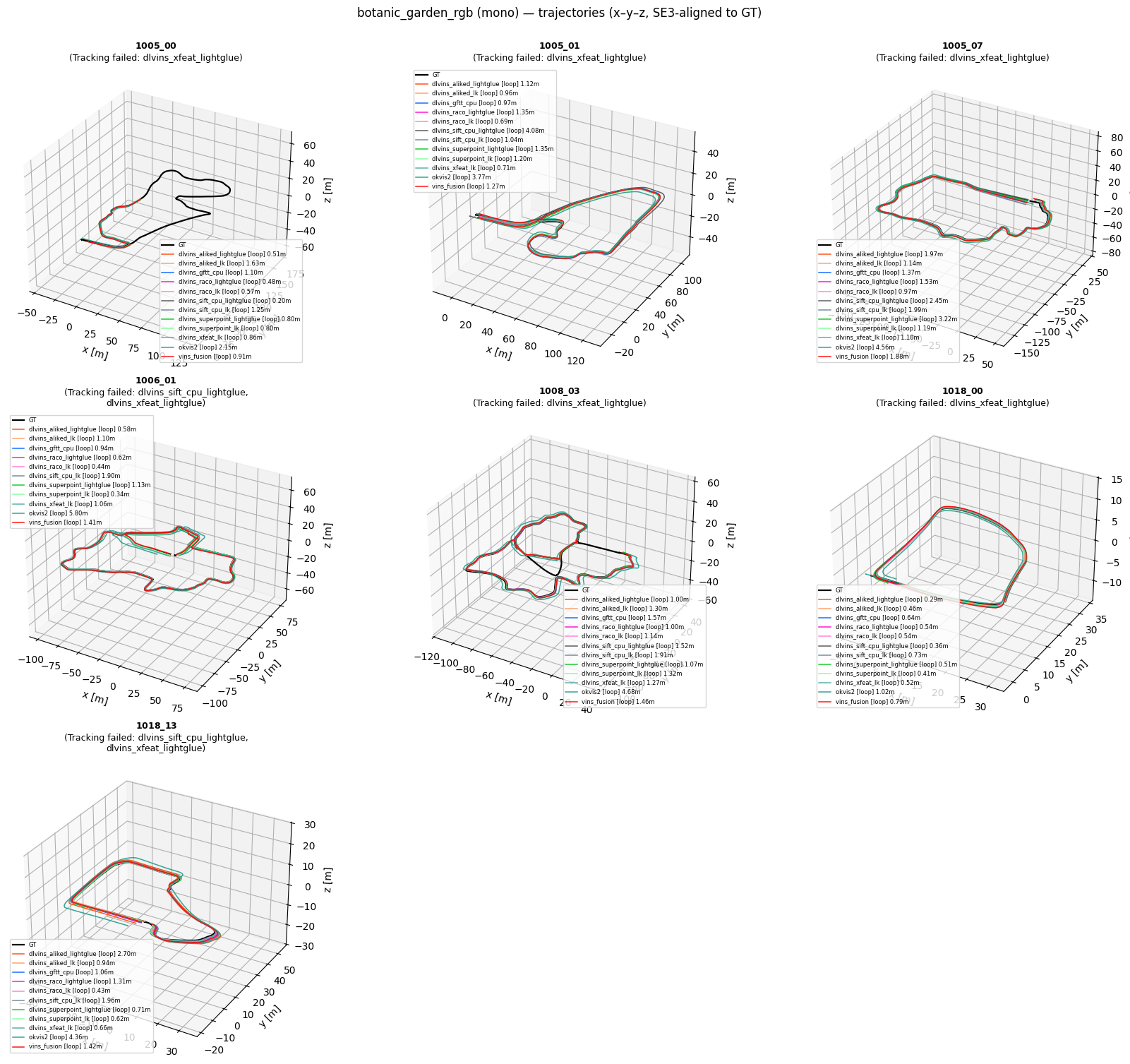}
  \caption{\textbf{Botanic Garden RGB} mono-inertial 3D trajectory plot.}
  \label{fig:bg_rgb_mono}
\end{figure*}

\begin{figure*}[t]
  \includegraphics[width=1.0\textwidth]{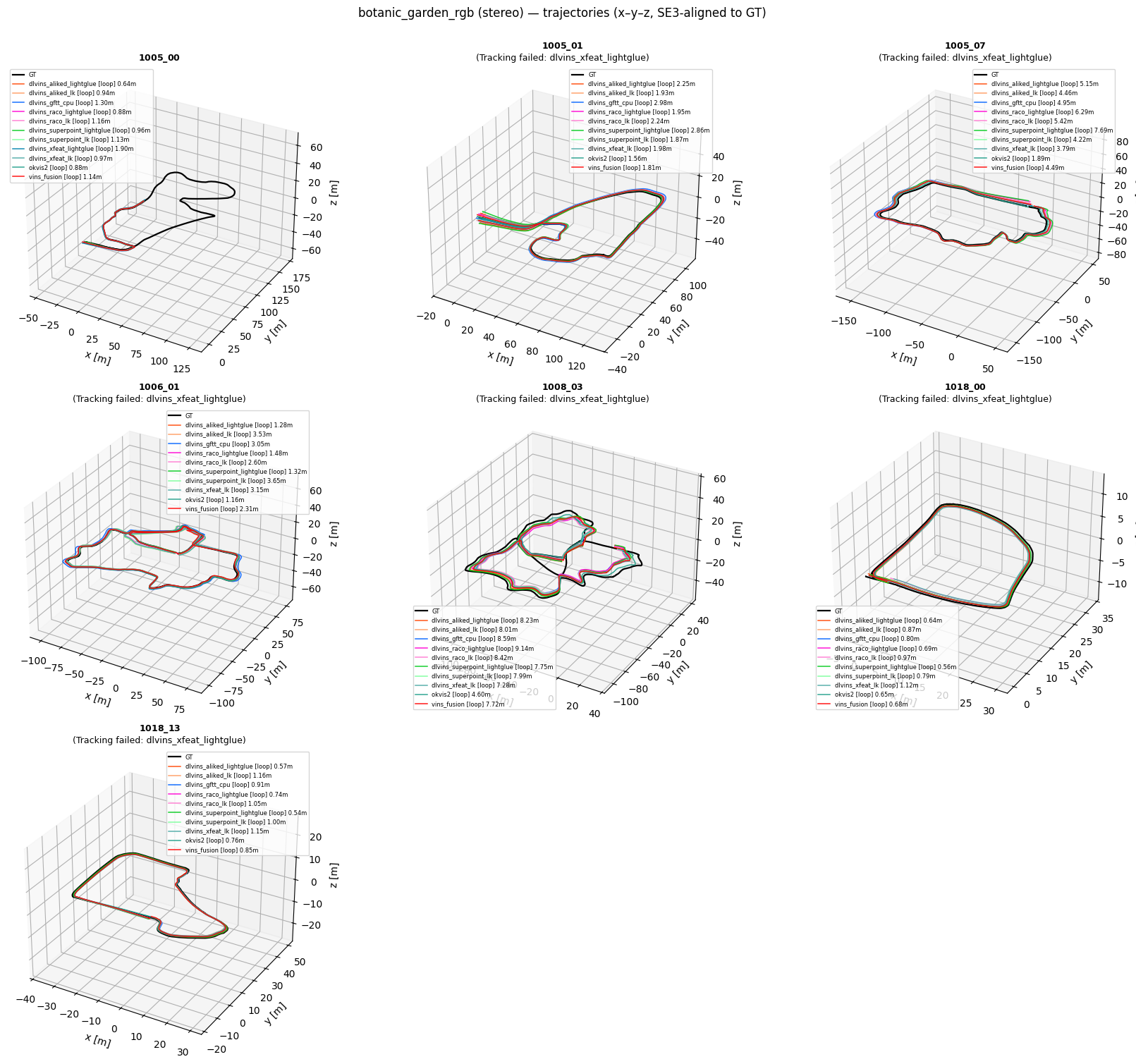}
  \caption{\textbf{Botanic Garden RGB} stereo-inertial 3D trajectory plot.}
  \label{fig:bg_rgb_stereo}
\end{figure*}

\begin{figure*}[t]
  \includegraphics[width=1.0\textwidth]{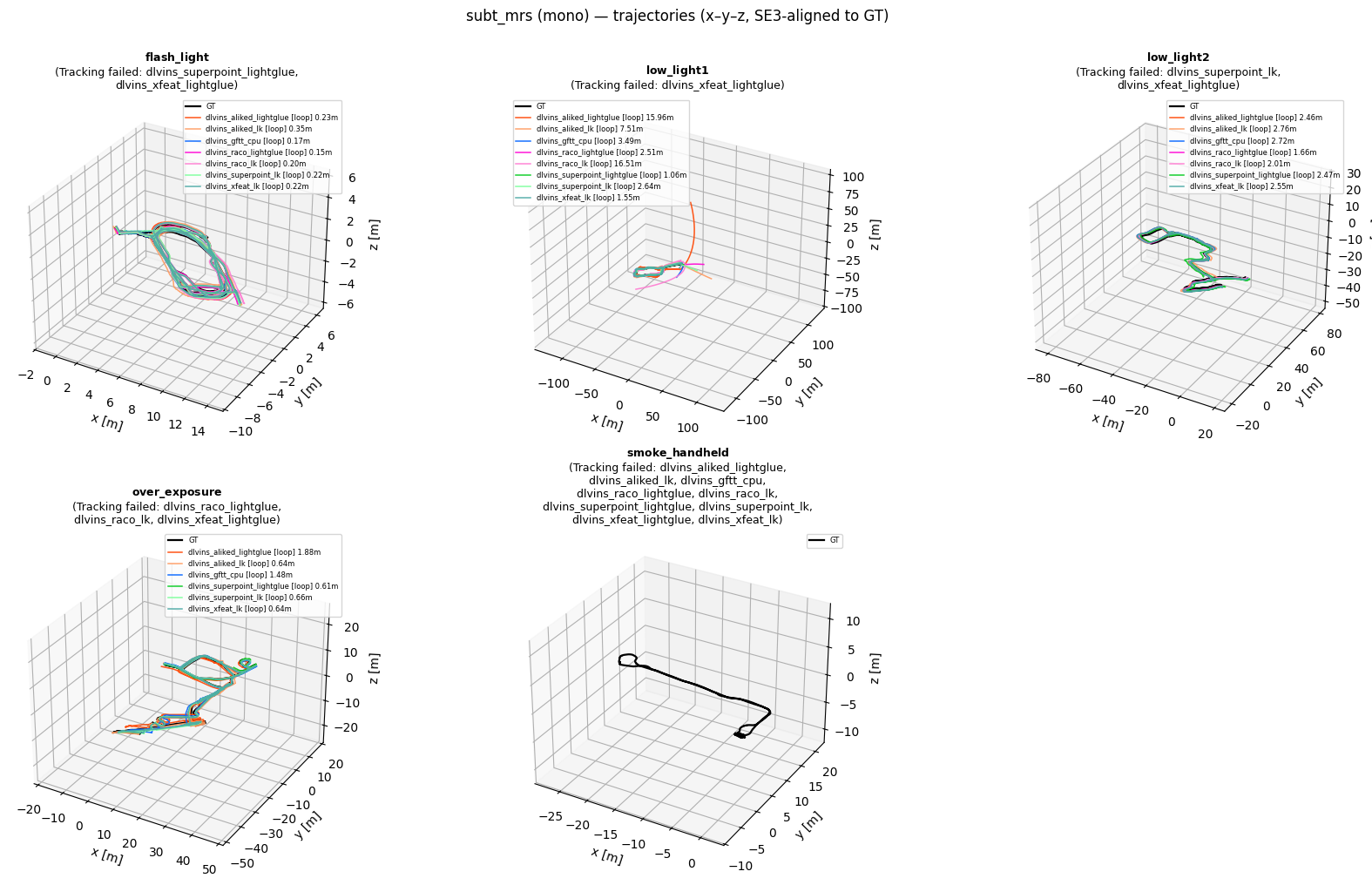}
  \caption{\textbf{SubT\_MRS} mono-inertial 3D trajectory plot.}
  \label{fig:subt_mrs}
\end{figure*}

\end{document}